\documentclass[lettersize,journal]{IEEEtran}
\usepackage{amsmath,amsfonts}
\usepackage{algorithmic}
\usepackage{algorithm}
\usepackage{array}
\usepackage[caption=false,font=normalsize,labelfont=sf,textfont=sf]{subfig}
\usepackage{textcomp}
\usepackage{stfloats}
\usepackage{url}
\usepackage{verbatim}
\usepackage{graphicx}
\usepackage{cite}
\usepackage{times}
\usepackage{microtype}
\usepackage{soul}
\usepackage[usenames,svgnames]{xcolor}
\usepackage[colorlinks,citecolor=black,linkcolor=black,urlcolor=black]{hyperref}
\usepackage[utf8]{inputenc}
\usepackage{setspace}
\usepackage[small]{caption}
\usepackage{graphicx}
\usepackage{amsmath}
\usepackage{amsthm}
\usepackage{booktabs}
\usepackage[switch]{lineno}
\usepackage{tabularx}
\usepackage{framed}
\usepackage{bbding}
\usepackage{subfloat}
\usepackage{adjustbox}
\usepackage{multirow}
\usepackage{enumitem}
\usepackage{mydef}
\usepackage{amssymb}
\usepackage{pifont}

\newcommand{\paratitle}[1]{\vspace{0.8ex}\noindent \textbf{#1}}
\newcommand{\figref}[1]{Figure~\ref{#1}}
\newcommand{\tabref}[1]{Table~\ref{#1}}

\definecolor{lightcoral}{rgb}{0.94, 0.5, 0.5}
\definecolor{lightgreen}{rgb}{0.56, 0.93, 0.56}
\definecolor{harvestgold}{rgb}{0.85, 0.57, 0.0}
\definecolor{brightlavender}{rgb}{0.75, 0.58, 0.89}
\definecolor{capri}{rgb}{0.0, 0.75, 1.0}
\definecolor{carminepink}{rgb}{0.92, 0.3, 0.26}
\definecolor{celadon}{rgb}{0.67, 0.88, 0.69}
\definecolor{darkpastelgreen}{rgb}{0.01, 0.75, 0.24}
\definecolor{darkgreen}{RGB}{150, 194, 145}

\definecolor{brightlavender}{RGB}{186, 148, 209}
\definecolor{lightcoral}{RGB}{230, 164, 180}
\definecolor{darkpastelgreen}{RGB}{150, 194, 145}
\definecolor{harvestgold}{rgb}{0.85, 0.57, 0.0}

\begin{document}


\title{A Survey of Cross-domain Graph Learning: Progress and Future Directions}

\author{Haihong~Zhao,~\IEEEmembership{}
        Zhixun~Li,~\IEEEmembership{}
        Chenyi~Zi,~\IEEEmembership{}
        Aochuan~Chen,~\IEEEmembership{}
        Fugee~Tsung,~\IEEEmembership{Senior Member,~IEEE}
        Jia~Li,~\IEEEmembership{}
        and Jeffrey~Xu~Yu,~\IEEEmembership{Senior Member,~IEEE}        
        \thanks{\noindent\rule[0.5ex]{0.45\linewidth}{1.0pt}}
        \thanks{(Corresponding~author: Jia~Li)}
        \IEEEcompsocitemizethanks{
        \IEEEcompsocthanksitem Haihong Zhao, Chenyi Zi, Aochuan Chen, Fugee Tsung, Jia Li, and Jeffrey Xu Yu are with The Hong Kong University of Science and Technology (Guangzhou), Guangzhou, China (e-mail: hzhaobf@connect.hkust-gz.edu.cn; ci447@connect.hkust-gz.edu.cn; achen149@connect.hkust-gz.edu.cn; season@ust.hk; jialee@ust.hk; jeffreyxuyu@hkust-gz.edu.cn).\protect
        \IEEEcompsocthanksitem Zhixun Li is with The Chinese University of Hong Kong, Hong Kong, China (e-mail: zxli@se.cuhk.edu.hk). \protect}
        }

\markboth{Journal of \LaTeX\ Class Files,~Vol.~14, No.~8, August~2021}%
{Shell \MakeLowercase{\textit{et al.}}: A Sample Article Using IEEEtran.cls for IEEE Journals}


\maketitle

\begin{abstract}
    Graph learning plays a vital role in mining and analyzing complex relationships within graph data and has been widely applied to real-world scenarios such as social, citation, and e-commerce networks. Foundation models in computer vision (CV) and natural language processing (NLP) have demonstrated remarkable cross-domain capabilities that are equally significant for graph data. However, existing graph learning approaches often struggle to generalize across domains. Motivated by recent advances in CV and NLP, cross-domain graph learning (CDGL) has gained renewed attention as a promising step toward realizing true graph foundation models. In this survey, we provide a comprehensive review and analysis of existing works on CDGL. We propose a new taxonomy that categorizes existing approaches according to the type of transferable knowledge learned across domains: structure-oriented, feature-oriented, and mixture-oriented. Based on this taxonomy, we systematically summarize representative methods in each category, discuss the key challenges and limitations of current studies, and outline promising directions for future research. A continuously updated collection of related works is available at: \url{https://github.com/cshhzhao/Awesome-Cross-Domain-Graph-Learning}.   
\end{abstract}

\section{Introduction}

\IEEEPARstart{G}{raph}, or graph theory, has been playing an increasingly crucial role across numerous fields, actively contributing to various industries in the modern world, including recommendation~\cite{zang2022relatedsurvey2, liu2020modelling}, finance~\cite{li2023digaBank1,deprez2024networkBank2}, weather~\cite{liu2025cirt}, pharmaceuticals~\cite{li2022Pharm1,li2023Pharm2molecularalign}, and communications~\cite{ji2022communication2,zhao2023communication1}. Graph data can model the structural characteristics between nodes, offering significant advantages in capturing interactions and dependencies across structures and features. Many real-world datasets (e.g., social, citation, molecular, e-commerce, and transportation networks~\cite{sen2008citation1,li2023Pharm2molecularalign,zhu2024graphclip,yu2024temporalEcommerce1,jin2024urbanTransport1}) are naturally represented as graphs. To efficiently process and analyze these complex graph data, Graph Neural Networks (GNNs)~\cite{kipf2016gcn,velivckovic2018gat,liusegno} have emerged as a leading technique. The primary goal of GNNs is to learn the relationships between nodes and their neighbors through recursive message passing and aggregation mechanisms, thereby obtaining expressive representations at the node, edge, or graph level for various downstream tasks.

\begin{figure}[t]
  \centering
  \includegraphics[width=0.485\textwidth]{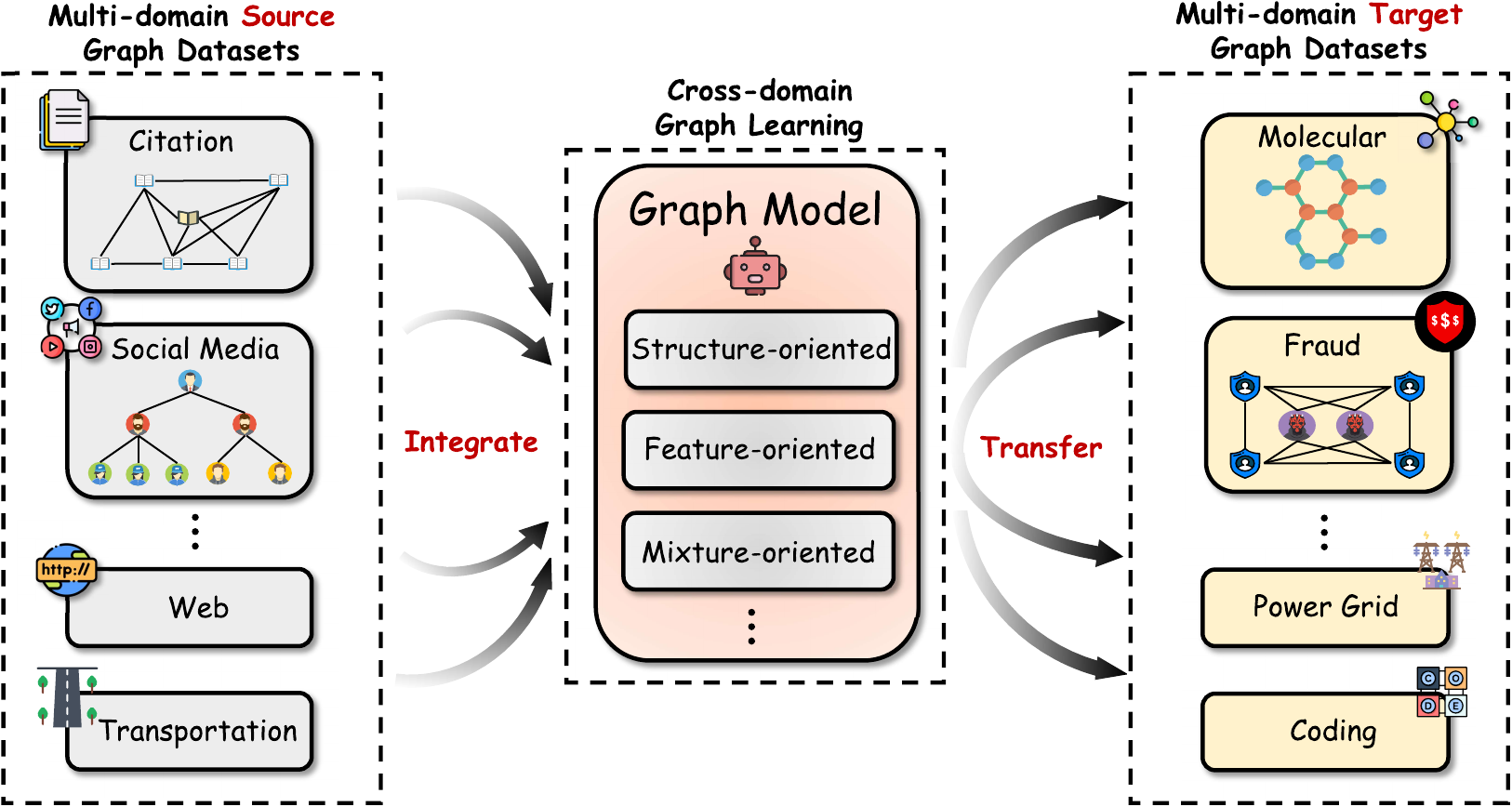}
  \caption{Cross-domain graph learning aims to integrate knowledge from multi-domain source graphs and transfer it to diverse target domains.}
  \label{fig:overview}
  \vspace{-2em}
\end{figure}

In recent years, the rapid development of foundation models, especially the large language models (LLMs)~\cite{chang2024llmsurvey1,min2023llmsurvey2}, has significantly revolutionized the way people live and work, such as ChatGPT in natural language processing (NLP)~\cite{achiam2023gpt4} and DALL-E 3 in computer vision (CV)~\cite{betker2023DALLE3}. Compared to previous deep learning models, the core advantage of foundation models lies in their ability to train on data from diverse domains, subsequently transferring the learned knowledge across different domains, encompassing three key capabilities: cross-modal, cross-domain, and cross-task~\cite{zhao2023LlmThreeCapabilities1,sun2023LlmThreeCapabilities2}. 
\textbf{These three capabilities are also crucial in graph domains.} For instance, \textbf{cross-modal} capabilities allow user-item interaction networks to combine user relationships (graph structure) with textual content (text modality) to more accurately predict user interests, thus enhancing recommendation systems~\cite{wu2025graphCrossModal}. \textbf{Cross-domain} capabilities facilitate knowledge transfer between different domains—such as leveraging insights from transportation networks to optimize power load management in power grid networks~\cite{huang2024towardCrossDomainPower1,yang2025recentCrossDomainPower2}, or using rich user interaction data from social networks to improve anomaly detection in banking transaction networks with limited data~\cite{li2023digaBank1}. \textbf{Cross-task} capabilities enable models trained for one task (e.g., node classification to predict a paper’s research field) to be directly applied to another task (e.g., link prediction to identify paper citations), thus enhancing adaptability and efficiency across tasks~\cite{sun2023crossTask1,wang2025GIT}. Therefore, realizing a graph foundation model (GFM) equipped with the three key capabilities is essential for advancing the field of graph learning.

\paratitle{The importance of cross-domain graph learning (CDGL).} 
To propel graph learning into the era of true foundation models, the top priority is to develop an effective method for distilling commonalities from cross-domain graph data, enabling the representation space of pretrained graph models to generalize effortlessly across diverse domains.
In detail, single-domain cross-modal graph learning (e.g., integrating images, text, and graphs) can benefit from advancements in CV and NLP, where many methods have successfully integrated multi-modal information~\cite{jin2024crossModalsurvey1}. Additionally, single-domain cross-task graph learning can be addressed using classical graph pretraining models~\cite{you2020crossTask2,sun2023crossTask1,zhao2024crossTasksurvey1}. However, when handling cross-domain graph learning (Figure~\ref{fig:overview}), the diversity and complexity of graph data across domains make it difficult for graph models to extract universal knowledge from source domains and transfer it to target domains~\cite{zhao2024GCOPE}. This challenge primarily manifests in two aspects:
\begin{itemize}[leftmargin=15pt,labelsep=0.5em, itemsep=0pt, topsep=0pt]
    \item [(a)] \textbf{Structural differences}: Graphs from different domains can vary significantly in both connectivity and scale. Some graphs are sparsely connected, such as citation networks like Cora~\cite{he2025cora}, where most papers cite only a few others. In contrast, graphs like those in COLLAB~\cite{yanardag2015COLLAB} are densely connected, as researchers often collaborate extensively within the same subfield. In terms of scale, taking Flickr~\cite{tang2009Flickr,gao2024flickr} as an instance, it forms a large-scale social network graph where users are nodes and “contact” links are edges, whereas molecular datasets consist of many small graphs~\cite{li2022molecularsmall}, each representing a single molecule. These differences constitute diverse structural patterns; each domain’s unique characteristic can trigger structure-level negative transfer~\cite{zhu2021structurelevelNegativeTransfer}.
    \item [(b)] \textbf{Feature differences}: Node features in graphs from different domains are domain-specific, resulting in differences in both dimensionality and semantics. For example, in citation networks, node features typically represent the title and abstract of a paper, resulting in high-dimensional node features~\cite{li2024zerog}. In contrast, in molecular graphs, node features describe properties of atoms (e.g., atom type, atomic mass, hybridization), which are fewer in number, leading to low-dimensional node features~\cite{li2022molecularsmall}. Semantic gaps hinder identifying cross-domain feature commonalities~\cite{wu2024GDomainAdaptationSurvey2}, while dimensional mismatches can even prevent graph models from learning across domains~\cite{zhao2024GCOPE}.
\end{itemize}

Taken together, these differences pose significant challenges for effectively learning from cross-domain graph knowledge and producing generalizable graph representations. 
\textbf{Drawing on the foregoing analysis, CDGL emerges as the pivotal research direction toward genuine GFMs.} By harnessing CDGL, we can distill shared structural and feature patterns across different graph domains, endowing learned representations with robust, domain-agnostic transferability.

\tikzstyle{leaf}=[draw=hiddendraw,
    rounded corners,
    minimum height=1em,
    fill=mygreen!40,
    text opacity=1, 
    align=center,
    fill opacity=.5,  
    text=black,
    align=left,
    font=\scriptsize,
    inner xsep=3pt,
    inner ysep=1pt,
    ]
\tikzstyle{middle}=[draw=hiddendraw,
    rounded corners,
    minimum height=1em,
    fill=output-white!40, 
    text opacity=1, 
    align=center,
    fill opacity=.5,  
    text=black,
    align=left,
    font=\scriptsize,
    inner xsep=3pt,
    inner ysep=1pt,
    ]
    
\begin{figure*}[ht]
\centering
\resizebox{0.98\textwidth}{!}{%
\begin{forest}
  for tree={
  forked edges,
  grow=east,
  reversed=true,
  anchor=base west,
  parent anchor=east,
  child anchor=west,
  base=middle,
  font=\scriptsize,
  rectangle,
  line width=0.7pt,
  draw=output-black,
  rounded corners,align=left,
  minimum width=2em,
    s sep=5pt,
    inner xsep=3pt,
    inner ysep=1pt,
  },
  where level=1{text width=4.5em}{},
  where level=2{text width=6em,font=\scriptsize}{},
  where level=3{font=\scriptsize}{},
  where level=4{font=\scriptsize}{},
  where level=5{font=\scriptsize}{},
  [Cross-domain Graph Learning,middle,rotate=90,anchor=north,edge=output-black
    [Structure-oriented, middle, edge=output-black, text width=5.0em
        [Generative, middle, text width=3.7em, edge=output-black
            [GraphControl \cite{zhu2024graphcontrol}{,} GA$^{2}$E \cite{hu2024GA2E}{,} OpenGraph \cite{xia2024opengraph}{,} UniAug \cite{tang2024UniAug}{,} GFT \cite{wang2024gft}{,} GFSE \cite{chen2025gfse}, leaf, text width=25.75em, edge=output-black]
        ]
        [Contrastive, middle, text width=3.7em, edge=output-black
            [GCC \cite{qiu2020gcc}{,}
            PCRec\cite{wang2021pcrec}{,} GRADE \cite{wu2023GRADE}{,} APT \cite{xu2023APT} {,} FedStar \cite{tan2023FedStar}{,} BooG \cite{cheng2024BooG}{,} \\ProCom \cite{wu2024procom}{,} RiemannGFM~\cite{sun2025RiemannGFM}, leaf, text width=25.75em, edge=output-black]
        ]
    ]
    [Feature-oriented, middle, edge=output-black, text width=5.0em
        [In-space, middle, text width=3.7em, edge=output-black
            [KTN \cite{yoon2022KTN}{,} NaP \cite{wang2024NAP} {,} GPF-plus~\cite{fang2024gpfplus}{,} RELIEF~\cite{zhu2024relief}{,} DAGPrompT~\cite{chen2025DAGPrompT}{,} GADT3~\cite{pirhayatifard2025crossdomainGADT3}, leaf, text width=25.75em, edge=output-black]
        ]
        [Cross-space, middle, text width=3.7em, edge=output-black
            [OFA~\cite{liu2024OFA} {,} GraphAlign \cite{hou2024graphalign}{,} CDFS-GAD \cite{chen2024CDFSGAD}{,} ZeroG~\cite{li2024zerog}{,} UNPrompt~\cite{niu2024UNPrompt}, leaf, text width=25.75em, edge=output-black]
        ]
    ]
    [Mixture-oriented, middle, edge=output-black, text width=5.0em
        [Sequential, middle, text width=3.7em, edge=output-black
            [PTGB \cite{yang2023ptgb}{,} SDGA \cite{qiao2023SDGA}{,} SOGA \cite{mao2024SOGA}{,} GSPT \cite{song2024GSPT}{,} ARC \cite{liu2024ARC}{,} UniGraph \cite{he2024unigraph}{,}\\ AnyGraph \cite{xia2024anygraph}{,} GraphLoRA~\cite{yang2024graphlora}{,} SAMGPT~\cite{yu2025SAMGPT}{,}
            MDGFM~\cite{wang2025MDGFM}{,}
            MLDGG~\cite{tian2025mldgg}, leaf, text width=25.75em, edge=output-black]
        ]
        [Unified, middle, text width=3.7em, edge=output-black
            [Graph-based, middle, text width=3.7em, edge=output-black
                [UDA-GCN \cite{wu2020UDAGCN}{,} DA-GCN\cite{Guo2021DAGCN}{,} COMMDER\cite{ding2021COMMANDER}{,} GCFL\cite{xie2021gcfl}{,}\\ CDGEncoder \cite{hassani2022CDGEncoder}{,} CrossHG-Meta \cite{zhang2022CrossHGMeta}{,} METABrainC \cite{yang2022METABrainC}{,}\\ CCDR \cite{xie2022CCDR}{,} GAST \cite{zhang2022GAST}{,} DH-GAT\cite{xu2023DHGAT}{,} ALEX\cite{yuan2023alex}{,} ACT \cite{wang2023ACT}{,}\\ CDTC \cite{zhang2023CDTC}{,} DGASN \cite{shen2023DGASN}{,} STGP \cite{hu2024stgp}{,} GCOPE \cite{zhao2024GCOPE}{,} ALCDR \cite{zhao2023ALCDR}{,}\\ PGPRec~\cite{yi2023PGPRec}{,} CrossLink\cite{huang2024CrossLink}{,} MDGPT\cite{yu2024MDGPT}{,} Uni-GLM \cite{fang2024uniglm}{,}\\ OMOG \cite{liu2024OMOG}{,} MDP-GNN \cite{lin2025MDPGNN}{,} UniGraph2 \cite{he2025unigraph2}{,} GIT \cite{wang2025GIT}{,}\\ CMPGNN~\cite{wang2025bridgingCMPGNN}{,} AEGOT-CDKT~\cite{wu2025crossAEGOTCDKT},leaf, text width=20.25em, edge=output-black]
            ]
            [Flatten-based, middle, text width=3.7em, edge=output-black
                [GIMLET~\cite{zhao2023gimlet}{,} GraphTranslator~\cite{zhang2024graphtranslator}{,} GraphGPT~\cite{tang2024graphgpt}{,} GITA \cite{wei2024gita}{,} \\InstructGraph \cite{wang2024instructgraph}{,} TEA-GLM \cite{wang2024TeaGLM}{,} GOFA~\cite{kong2025gofa}{,} HiGPT \cite{tang2024higpt}{,}\\ GraphCLIP~\cite{zhu2024graphclip}{,} GraphWiz \cite{chen2024graphwiz}, leaf, text width=20.25em, edge=output-black]
            ]
        ]
    ]
  ]
\end{forest}
}
\caption{A taxonomy of graph models for solving cross-domain graph learning with representative examples.}
\label{fig:taxonomy_of_CDGLs}
\vspace{-1.5em}
\end{figure*}

\paratitle{Motivations.} Although CDGL has received increasing attention, this rapidly expanding field still lacks a systematic and comprehensive review. Concretely, some existing surveys have focused on graph domain adaptation or domain generalization methods, which primarily address the distribution shift between source and target domains~\cite{shi2024GDomainAdaptationSurvey1,wu2024GDomainAdaptationSurvey2,cai2024GDomainAdaptationSurvey3}. These approaches often assume that the domains share the same feature space, meaning the node features are dimensionally and semantically aligned. Additionally, certain CDGL methods are capable of transferring across domains, but these domains typically can be further attributed to the same high-level domain, such as cross-domain graph recommendation~\cite{zang2022relatedsurvey2,chen2024relatedsurvey1}. Beyond these methods, many recent methods have proposed novel training or evaluation strategies aiming to achieve the true cross-domain generalization required by foundation models~\cite{zhao2024GCOPE, niu2024UNPrompt, liu2024OMOG,sun2025RiemannGFM,chen2025DAGPrompT}. While each of these CDGL methods contributes valuable insights toward building real GFMs, they have largely evolved in isolation. The lack of systematic analysis among them hinders researchers from gaining a holistic understanding of the CDGL landscape and identifying future research.

To fill this gap, we aim to provide a systematic survey on CDGL, helping researchers better understand the current state of research and its challenges. We review representative CDGL works to organize our first-level taxonomy, as shown in Figure~\ref{fig:taxonomy_of_CDGLs}, which includes structure-oriented, feature-oriented, and mixture-oriented approaches. We further refine this taxonomy by introducing more granularity to the initial categories.

\paratitle{Contributions.} The contributions of this work can be summarized in the following three aspects. \textbf{(1)} \textit{A structured taxonomy}. A broad overview of the field is presented with a structured taxonomy that categorizes existing works into three categories (\figref{fig:taxonomy_of_CDGLs}). \textbf{(2)} \textit{A comprehensive review}. Based on the proposed taxonomy, the current research progress of cross-domain techniques for graph learning is systematically delineated. To the best of our knowledge, this is the first comprehensive survey for CDGL. \textbf{(3)} \textit{Some future directions}. We discuss the remaining limitations of existing works and point out possible future directions.
\section{Preliminary}

In this section, we first give the related notations and then introduce the basic concepts of two key areas related to this survey, i.e., graph representation learning and cross-domain learning. Finally, we give a brief introduction to the newly proposed taxonomy for CDGL.

\subsection{Notations}
Consider a graph instance represented as $\mathcal{G}=\{\mathcal{V},\mathcal{E}\}$, where $\mathcal{V}=\{v_1, v_2, \ldots, v_{N_v}\}$ is the set of $N_v$ nodes. The set $\mathcal{E}\subseteq \mathcal{V} \times \mathcal{V}$ defines the connections between nodes. Associated with each node $v_i$ is a feature vector $x_i \in \mathbb{R}^D$, and the node feature matrix is denoted as $\mathcal{X} \in \mathbb{R}^{N_v \times D}$. To describe the connectivity within the graph, we use the adjacency matrix $\mathcal{A} \in \{0, 1\}^{N_v \times N_v}$, where the matrix entry $\mathcal{A}_{i,j}$ is defined as 1 if and only if there is an edge $(v_i, v_j) \in \mathcal{E}$. In this paper, the graphs from the same domain lie in the same graph domain space, denoted as $\mathcal{D}_G = \{ (\mathcal{G}_i, \mathcal{L}_i) \}_{i=1}^{N_G}$, where each $\mathcal{G}_i$ is a graph, $\mathcal{L}_i$ denotes the corresponding label and $N_G$ is the number of graphs.

\subsection{Graph Representation Learning}

\paratitle{Definitions.} Graph representation learning is a powerful technique for managing complex and heterogeneous non-Euclidean graph data. It focuses on learning meaningful representations for nodes, edges, or entire graphs by mapping them into continuous vector spaces. This approach preserves the intrinsic structural and feature information of the graph, which can facilitate various tasks such as node classification, link prediction, and graph classification. These methods can be broadly categorized into two main branches: shallow embedding methods and deep graph neural networks (GNNs).

\paratitle{Shallow Embedding Methods.} Shallow embedding methods, such as node2vec~\cite{grover2016node2vec} and DeepWalk~\cite{perozzi2014deepwalk}, aim to map nodes into low-dimensional embeddings by preserving the network's similarity structure. These methods utilize techniques like factorization-based methods or random walks to achieve this goal. While providing flexibility for various downstream tasks, shallow embedding methods face limitations: they cannot generate embeddings for unseen nodes and lack the capability to incorporate node features effectively.

\paratitle{Deep Graph Neural Networks (GNNs).} Deep GNNs, including Graph Convolutional Networks (GCN)~\cite{kipf2016gcn} and Graph Attention Networks (GAT)~\cite{velivckovic2018gat}, maintain input node features as constants and optimize model parameters for specific tasks. These networks typically employ a message-passing mechanism, where nodes iteratively exchange and aggregate information with their neighbors. This process leads to more expressive graph representations that can effectively handle more complex graph-based tasks. In detail, the forward process of the message-passing mechanism can be defined as:

\[
    h_i^{(l)} = \mathbf{U} \left( h_i^{(l-1)}, f_{MP}(\{ h_i^{(l-1)}, h_j^{(l-1)} \mid v_j \in \mathcal{N}_i \}) \right)
\]
where $h_i^{(l)}$ is the feature vector of node $v_i$ in the $l$-th layer, and $\mathcal{N}_i$ is a set of neighbor nodes of node $v_i$. $f_{MP}$ denotes the message passing function of aggregating neighbor information, $\mathbf{U}$ denotes the update function with central node feature and neighbor node features as input. By stacking multiple layers, GNNs can aggregate messages from higher-order neighbors.

The prevalent methods in graph representation learning are currently based on GNN architectures due to their enhanced ability to capture complex relational patterns in graph data. However, both shallow embedding methods and GNNs are primarily designed to learn representations within a specific graph domain, limiting their effectiveness across different domains. This issue underscores the need for research into methods that can generalize across diverse graph-based applications, aiming to enhance the learning capacity of graph representation techniques to operate effectively across varying domains.

\subsection{Cross-domain Learning}
\label{sec:cross-domain learning}

Before formally introducing CDGL, we first provide an overview of cross-domain learning, a pivotal research direction in deep learning that plays a crucial role in the development of foundational models such as LLMs. Specifically, we present a unified definition of cross-domain learning and summarize common cross-domain strategies, laying the groundwork for a better understanding of CDGL.

\paratitle{Definitions.} Given $m$ source domains $\mathbb{D}_S = \{ \mathcal{D}_{S_1}, \dots, \mathcal{D}_{S_m} \}$ and $n$ target domains $\mathbb{D}_T = \{ \mathcal{D}_{T_1}, \dots, \mathcal{D}_{T_n} \}$, the objective of cross-domain learning is to learn a shared embedding space from the data in $\mathbb{D}_S$, and then generalize this learned embedding space to the target domains for various downstream tasks. Each domain $\mathcal{D}$ is defined as a dataset of labeled samples $\mathcal{D} = \{ (\mathbf{x}_i, y_i) \}_{i=1}^{N_D}$, where $\mathbf{x}_i$ denotes an input sample, $y_i$ is the corresponding label, and $N_D$ is the number of samples in the domain. Each dataset is referred to as a distinct domain, where a domain is characterized by its own data distribution and task-specific context.

First, we define the model $\mathcal{M}$ as a source encoder that maps each sample $\mathbf{x}_{S_i} \in \mathcal{D}_{S_i} \subseteq \mathbb{D}_S$ to a unified embedding space $E$, which can be also generalized to target domains:
\begin{equation}
    \mathcal{M} : \mathbf{x}_{S_i} \rightarrow \mathbf{e} \in E
\end{equation}

Next, we define the set of $n$ task-specific functions for target domains $\boldsymbol{\mathcal{F}} = \{ f_{T_1}, \dots, f_{T_n} \}$, where each \( f_{T_j} \) aims to map the embeddings $\mathcal{M}(\mathbf{x}_{T_j})$ to the corresponding task result $y_{T_j}$:
\begin{equation}
    \boldsymbol{\mathcal{F}} = \{ f_{T_j}: \mathcal{M}(\mathbf{x}_{T_j}) \rightarrow y_{T_j}, \ \text{for} \ j = 1, \dots, n \}.
\end{equation}

Finally, we train the model $\mathcal{M}$ and the task-specific functions $\boldsymbol{\mathcal{F}}$ such that the features across all domains are unified into a shared embedding space, allowing effective cross-domain knowledge transfer, which in turn improves generalization and performance on the target tasks.

The objective function for training $\mathcal{M}$ can be defined as:
\begin{equation}
    \min_{\mathcal{M}} \sum_{i=1}^{m} \mathbb{E}_{(\mathbf{x},y) \sim P_{S_i}(\mathbf{x},y)} \left[ L_\mathcal{M}(\mathcal{M}(\mathbf{x}), y) \right],
\end{equation}
where $L_\mathcal{M}$ can be an unsupervised or supervised loss function (e.g., contrastive loss or classification loss), and $P_{S_i}(\mathbf{x}, y)$ represents the joint distribution of the source domain data.

The objective function for learning the set of task functions $F$ can be defined as:
\begin{equation}
    \min_{f_{T_j}} \sum_{j=1}^{n} \mathbb{E}_{(\mathbf{x},y) \sim P_{T_j}(\mathbf{x},y)} \left[ L_T(f_{T_j}(\mathcal{M}(\mathbf{x})), y) \right],
\end{equation}
where $L_T$ is the appropriate loss function for the target domain tasks.

Note that the two processes can be conducted either jointly or sequentially, depending on the learning paradigm (e.g., domain adaptation, pre-train \& fine-tune, or pre-train \& inference). For instance, domain adaptation typically performs joint training of $\mathcal{M}$ and $\boldsymbol{\mathcal{F}}$, while pre-train \& fine-tune follows a sequential pipeline, where the pre-training stage is typically unsupervised and does not rely on label supervision from $\mathbb{D}_S$ in the first process. In contrast, in pre-train \& inference (e.g., zero-shot learning), $\mathcal{M}$ trained in the first stage is directly applied to downstream tasks via embedding similarity without additional task-specific training. The degree of domain overlap $|\mathbb{D}_S \cap \mathbb{D}_T|$ typically reflects, to some extent, the difficulty of transferring cross-domain knowledge during the training process.

\paratitle{The development of cross-domain learning in various fields.} To better understand this research direction, we introduce the development of cross-domain learning in the fields of Natural Language Processing (NLP) and Computer Vision (CV), focusing on how cross-domain learning is achieved within these domains.
\begin{itemize}[leftmargin=*]
    \item[1] \textbf{NLP.} The development of cross-domain learning in NLP has progressed significantly, driven by innovations in model architectures, training strategies, and data availability. Early efforts relied on domain adaptation techniques, such as feature alignment and adversarial learning, to transfer knowledge from resource-rich domains to low-resource ones. Alongside these, domain generalization techniques, which aim to improve model robustness by learning domain-invariant representations, have also been widely explored. The rise of pre-trained language models (PLMs), like BERT~\cite{devlin2019bert} and GPT~\cite{floridi2020gpt3}, enabled generalization across domains by leveraging contextual embeddings and task-specific fine-tuning. More recently, large language models (LLMs), such as GPT-3~\cite{floridi2020gpt3} and GPT-4~\cite{achiam2023gpt4}, have demonstrated strong cross-domain capabilities through massive pretraining on diverse corpora, excelling in zero-shot and few-shot learning. Furthermore, multi-modal techniques have extended cross-domain learning beyond text~\cite{zhuminigptMultimodalGPT1,chen2024mllmMultimodalGPT2}, allowing models to integrate information from other modalities, such as images and structured data, to address complex tasks like visual question answering~\cite{guo2023visualquestion1} and multi-modal dialogue~\cite{wu2024visionllmMultimodalDialogue}.
    \item[2] \textbf{CV.} The development of cross-domain learning in computer vision (CV) follows a similar trajectory to that of NLP~\cite{xu2025deepSurveyCV1, wang2023visionfoundation1,liu2024vmambavisionfoundation2}. Early approaches rely on domain adaptation techniques~\cite{venkateswara2017DomainAdaptationCV1} such as adversarial training, feature alignment, and style transfer to address challenges posed by domains with different visual characteristics. As the field advances, domain generalization methods~\cite{wang2022DomainGeneralizationCV1}, including invariant feature learning and meta-learning, emerge to improve model robustness to unseen domains. The introduction of pre-trained vision models like ResNet~\cite{he2016deepResNet} and Vision Transformers (ViTs)~\cite{dosovitskiy2020ViT} further the progress by enabling models to generalize across domains through the reuse of learned features, which can then be fine-tuned for downstream tasks. Recently, multi-modal models like CLIP~\cite{zhu2024graphclip} and Sora~\cite{liu2024sora} demonstrate exceptional cross-domain capabilities by integrating visual and language understanding. For instance, Sora excels in tasks like image captioning, visual question answering, and zero-shot reasoning, benefiting from large-scale pretraining on multi-modal datasets. Just as in NLP, the combination of visual and textual data allows these models to achieve superior cross-domain performance~\cite{xuadvancingMultimodalCrossdomain1,xiang2025MultimodalCrossdomain2}.
    \item[3] \textbf{Summary.} The rapid development of cross-domain learning in NLP and CV can be attributed to the relatively straightforward alignment of features across domains, such as text and images. These fields benefit from shared representation formats—text and images—where semantic or visual features are easier to align across different domains. As a result, when the data scale increases, models can more easily learn shared cross-domain patterns, enabling better generalization.
\end{itemize}

In contrast, graph data presents unique challenges for cross-domain learning. Unlike text or images, graphs are not inherently structured and require manual definition of their structures, leading to complexity in graph construction and a lack of uniformity in graph representations. Additionally, graph features often vary widely across domains, and some features may not even carry semantic information, making them difficult to align. This inherent complexity and diversity in graph data contribute to the slower development of cross-domain learning in the graph domain, which remains an area of active research.

\subsection{Cross-domain graph learning}
\label{sec:cross-domain graph learning}
CDGL is a key yet underexplored area within cross-domain learning. Building on the previous definition, CDGL can be described by specifying the $X_{T_i}$ using graph data $\mathcal{G}_{T_i}$.

\paratitle{Definitions.} Derived from cross-domain learning, CDGL aims to capture cross-domain knowledge from given $m$ source graph domains $\mathbb{D}_S = \{\mathcal{D}_{G_{S_1}},\dots,\mathcal{D}_{G_{S_i}}|i=1,\dots,m \}$ and then transfer the captured knowledge to learn a set of task functions (e.g., node classification, graph classification, etc.) on $n$ target graph domains $\mathbb{D}_T = \{\mathcal{D}_{G_{T_1}},\dots,\mathcal{D}_{G_{T_i}}|i=1,\dots,n \}$.

\subsubsection{The Refinement of Cross-domain Scale} 
The source and target domains in cross-domain graph learning (CDGL) are typically defined manually, and this definition plays a crucial role in determining the scope of cross-domain learning. The manner in which these domains are defined influences the scale, which in turn reflects the complexity and diversity of the domains involved. In this survey, we categorize CDGL into three distinct scales, as described below:

\begin{itemize}[leftmargin=*]
    \item[1] \textbf{Limited cross-domain graph learning} refers to scenarios where graph samples from source and target domains share a set of similar or consistent attributes (i.e., from the same field). For example, NaP~\cite{wang2024NAP} divides the Facebook100 dataset—which contains social networks from 100 U.S. universities—into different domains, treating the social network of each university as a unique domain. The data from each domain share similar attributes, except for the university types. CDGL is then conducted across different universities. Similarly, SDGA~\cite{qiao2023SDGA} treats ACMv9, Citationv1, and DBLPv7—datasets constructed from ArnetMiner~\cite{tang2008Arnetminer}—as distinct domains for cross-domain learning.
    \item[2] \textbf{Conditional cross-domain graph learning} refers to situations where the boundaries between source domains and graph domains are clearly defined, but their attributes can be transformed into a common format (e.g., text) and covered by a specific high-level field. For example, OFA~\cite{liu2024OFA} can convert cross-domain graph data into a text-attribute graph (TAG) format. However, it faces difficulties when applied to non-TAG domains, such as brain networks~\cite{yang2022METABrainC}.
    \item[3] \textbf{Open cross-domain graph learning} refers to scenarios where graph data from both source and target domains can be arbitrary and do not need to satisfy specific conditions. For example, GCC~\cite{qiu2020gcc} focuses on leveraging only the structural information from datasets across domains, using it to pre-train models and learn common structural patterns in graph data. GCOPE~\cite{zhao2024GCOPE} presents a general framework for cross-domain graph pre-training that unifies the feature dimensions of graph data with distinct boundaries, enabling pre-training across domains for better generalization. Importantly, these unified graph features can represent data from any domain.
\end{itemize}

In general, from the perspective of node features, if a CDGL method requires that the feature dimensions across datasets remain consistent, it typically falls under the limited scale; if it relies on mapping features into an existing shared representation space (e.g., a text space), it is conditional; and if it can align inherently diverse feature dimensions for cross-domain knowledge transfer, it belongs to the open scale.

\subsubsection{The Refinement of Cross-domain Difficulty}

Beyond cross-domain scales, we further discuss the difficulty of learning cross-domain knowledge during training (i.e., cross-domain difficulty) to better understand CDGL. Within the same cross-domain scale, different methods encounter varying levels of difficulty due to differences in the overlap between source and target domains. Consequently, we delineate cross-domain difficulty into three distinct levels based on the magnitude of the intersection between the source and target domains:

\begin{itemize}[leftmargin=*]
\item[1] \textbf{High-Difficulty (When \( |\mathbb{D}_T \cap \mathbb{D}_S| = 0 \))}: In this case, the source and target domains share no common data or features, indicating a complete lack of overlap. This extreme training setting is often adopted when the target domain has very limited data. It also aligns with the ultimate goal of CDGL—achieving open cross-domain graph learning under challenging conditions. A model that performs well here typically demonstrates strong generalizability.

\item[2] \textbf{Moderate-Difficulty (When \( 0< |\mathbb{D}_T| - |\mathbb{D}_T \cap \mathbb{D}_S| < |\mathbb{D}_T| \))}: In this scenario, the target domain shares either partial or complete overlap with the source domain. The difficulty is moderate because CDGL models can learn shared cross-domain knowledge in advance, reducing the overall challenge. However, unique domain characteristics may still require fine-tuning or domain-specific adjustments to address discrepancies.

\item[3] \textbf{Low-Difficulty (When \( |\mathbb{D}_T| - |\mathbb{D}_T \cap \mathbb{D}_S| =  0 \))}: This scenario indicates that the target domain is entirely equal to the source domain, with both domains sharing the same representation space. Cross-domain learning in this setting is relatively easy because the model does not need to transfer knowledge from source to target domains but only needs to capture cross-domain knowledge directly. This situation is more suitable for cases where both domains have abundant data, and the goal is to use a single model to handle tasks across different domains. Note that, in general, high overlap does not necessarily guarantee good generalization.
\end{itemize}

By addressing these levels, we can better evaluate and develop models for diverse scenarios. Note that some approaches (e.g., domain adaptation approaches), which explicitly delineate source and target domains and ensure that the datasets do not overlap, actually involve both labeled source and unlabeled target samples in training. In such cases, we should consider the source and target domains as completely overlapping, representing a low-difficulty scenario. Ultimately, achieving robust performance in high-difficulty settings is key to advancing open cross-domain graph learning.

\subsubsection{Proposed Taxonomy} Considering the key elements of graph data (including nodes, edges, adjacency matrix, and node features, which collectively encompass both structural and feature information) and referring to the definition of the given CDGL, we propose a taxonomy (as illustrated in \figref{fig:taxonomy_of_CDGLs}) that organizes representative techniques focused on capturing generalizable cross-domain information into three main categories: \textbf{(1)} Structure-oriented approach, where the graph structural information from various domains is utilized as the basis for cross-domain learning. \textbf{(2)} Feature-oriented approach, where the graph feature information from various domains is specifically learned and leveraged. \textbf{(3)} Mixture-oriented approach, where both the graph structural and feature information from various domains are organically integrated and analyzed from a unified perspective.

In the following sections, we present a comprehensive survey along the three main categories of our taxonomy for learning cross-domain graph knowledge in graph-related tasks, respectively.
\section{Structure-oriented Approach}
\label{sec:structure}
Structural information, as the distinctive and core characteristic of graph data, is essential for addressing real-world graph-related problems. Existing graph models have demonstrated the ability to effectively capture structural information in single-domain scenarios. However, in cross-domain scenarios, the graphs from different domains exhibit varying node connectivity patterns, with some being densely connected and others more sparsely connected. These structural differences pose significant challenges to current graph models. To address this, structure-oriented approaches aim to improve the performance of existing models by extracting commonalities in structural information across domains. Based on how structural information is utilized, these approaches can be broadly categorized into two main branches: Generative and Contrastive.

\subsection{Generative}

\begin{figure}[t]
  \centering
  \includegraphics[width=0.43\textwidth]{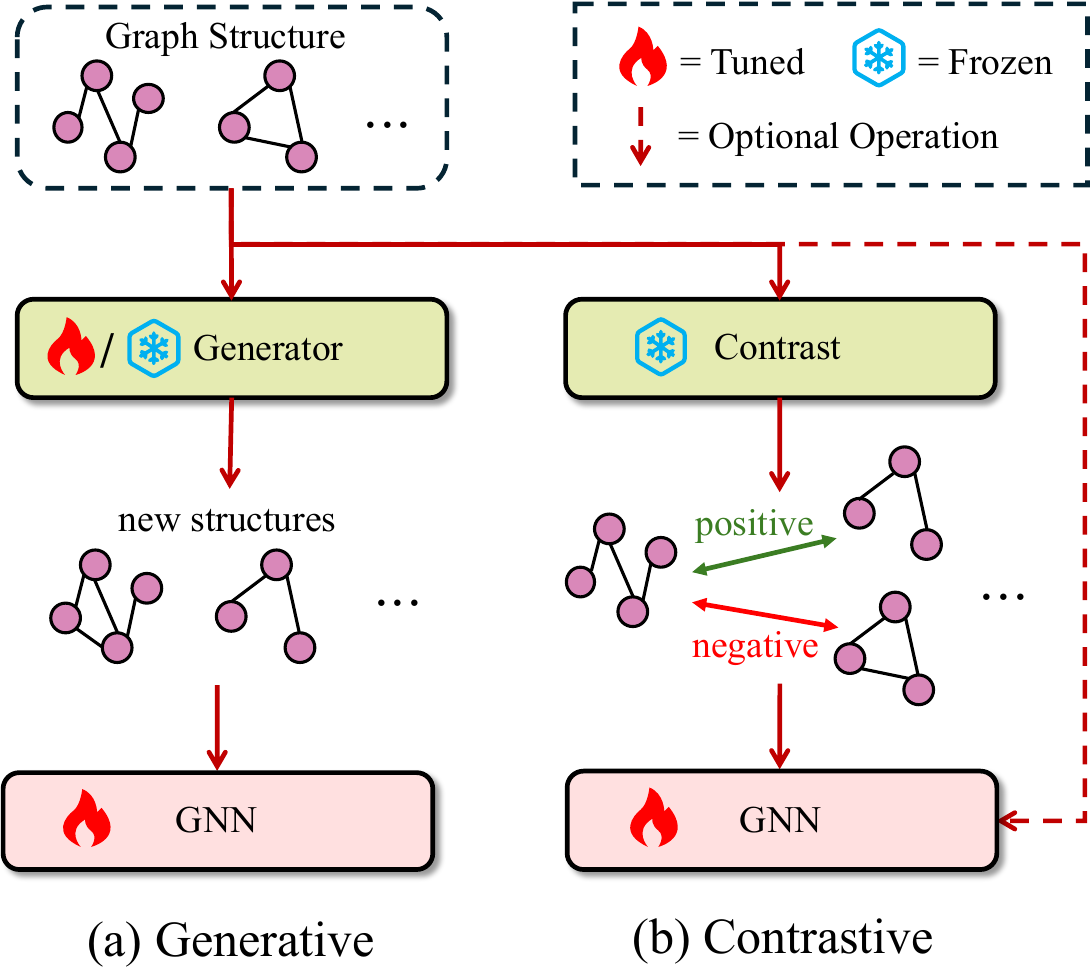}
  \caption{Structure-oriented CDGL: (a) a shared generator produces diverse structures for both source and target domains; (b) structure contrast explicitly or implicitly constructs positive–negative pairs in the source domain to improve generalization to target domains.}
  \label{fig:overview of structure-oriented approaches}
\end{figure}

To effectively capture structural information, a useful approach is to establish a unified node connection strategy for graphs, ensuring that the graph data in both source domains and target domains follows the same connection rules. This allows the cross-domain graph data to share the same structural space. Structure-oriented generative CDGL methods are developed based on this principle. It employs a shared module to generate new and distinct structures for graph data from both source and target domains. These newly generated structures enable the graph model to adopt a unified perspective for analyzing structural information across various domains, thereby enhancing the model's performance on different graph-related tasks. As shown in Figure~\ref{fig:overview of structure-oriented approaches}, generative methods typically begin by employing a structure generation module $Generator(\cdot)$ to edit and transform the input graph $\mathcal{G_S} \in \mathbb{D}_{G_S}$, resulting in new graph structures and thereby forming a modified source domain $\mathbb{D}^{'}_{G_S}$. A graph model $\mathcal{M}_{\mathbb{D}^{'}_{G_S}}$ is then trained on $\mathbb{D}^{'}_{G_S}$ and subsequently generalized to the target domain $\mathbb{D}_T$ via a task-specific prediction head $TaskH(\cdot)$, which outputs the final prediction results:
\begin{align}
    \text{Structure Generation:} \quad & \mathcal{G}^{'}_S = Generator(\mathcal{V}_S,\mathcal{E}_S,\mathcal{A}_S), \\
    \text{Generalization:} \quad & \hat{y}_T = TaskH(\mathcal{M}_{\mathbb{D}^{'}_{G_S}}(\mathcal{G}^{'}_T)),
\end{align}
Here, $Generator(\cdot)$ aims to generate unified structural patterns across domains, and $\mathcal{M}_{\mathbb{D}^{'}_{G_S}}$ is expected to generalize to $\mathbb{D}_T$ despite domain-specific differences.

For instance, GraphControl~\cite{zhu2024graphcontrol} utilizes structure pretraining on source graphs to learn common structural patterns. Inspired by ControlNet, it generates a new structure from target node features and combines it with the original graph structure. This feature-derived structure is injected into a trainable copy of the pretrained model through zero-MLPs, allowing the model to adapt to the target domain and effectively transfer structural knowledge.
UniAug~\cite{tang2024UniAug} introduces diffusion models into cross-domain graph learning for the first time. It pre-trains a graph diffusion model on source domain structures to learn shared structural patterns. Then, it modifies the target domain graphs using the frozen pre-trained diffusion model to generate new structures, which are used for fine-tuning or prompting a downstream classifier, thereby achieving structure-level transfer across domains. GA$^{2}$E~\cite{hu2024GA2E} is the first to reformulate graphs from different domains into subgraphs for cross-domain learning. It leverages an autoencoder framework during pretraining to embed and reconstruct these subgraphs, allowing the model to capture and unify diverse structural patterns. This subgraph-based reconstruction enables effective structural knowledge transfer to target domains. OpenGraph~\cite{xia2024opengraph} first designs a node tokenization scheme that converts nodes from various source domain graphs into a unified token set based on their structural information (i.e., adjacency matrices). A scalable Transformer is then pre-trained on these tokens to learn a shared representation space. To mitigate limited structural coverage in the training data, OpenGraph uses LLMs to synthesize entirely new node–edge sets, broadening the structural distribution. The pre-trained Transformer captures shared structural patterns that can be effectively transferred to target domains.
GFT~\cite{wang2024gft} transforms graphs into computation trees derived from GNN message passing, then generates a discrete tree vocabulary through self-supervised learning, embedding structural commonalities across domains. During fine-tuning, downstream tasks are reformulated as tree classification using this shared vocabulary, enabling efficient cross-domain transfer. GFSE~\cite{chen2025gfse} introduces four structure-aware self-supervised tasks (e.g., Shortest Path Distance Regression~\cite{Dijkstra1959shortestpath1,li2020shortestpath2}) during pre-training to uncover structural commonalities across multi-domain graph datasets. In particular, the Shortest Path Distance Regression task dynamically extracts and models diverse shortest-path substructures during the pre-training phase, thereby enabling the pre-trained model to understand cross-domain graph structures.

\subsection{Contrastive}

In contrast to structure generation, structure-oriented contrastive methods aim to compare structural patterns across different graph datasets and identify domain-invariant commonalities among them. As shown in Figure~\ref{fig:overview of structure-oriented approaches}, contrastive methods typically begin by employing a structure contrastive module $Contrast(\cdot)$ to compare and modify the input graph $\mathcal{G_S} \in \mathbb{D}_{G_S}$, which can produce positive and negative sample pairs $(\mathcal{G}^{pos}_S,\mathcal{G}^{neg}_S)$. A graph model $\mathcal{M}_{G_S}$ is then trained on the produced sample pairs and subsequently generalized to the target domain $\mathbb{D}_T$ via a task-specific prediction head $TaskH(\cdot)$, which outputs the final prediction results:
\begin{align}
    \text{Structure Contrast:} \quad & (\mathcal{G}^{pos}_S,\mathcal{G}^{neg}_S) = Contrast(\mathcal{V}_S,\mathcal{E}_S,\mathcal{A}_S), \\
    \text{Generalization:} \quad & \hat{y}_T = TaskH(\mathcal{M}_{G_S}(\mathcal{G}_T)),,
\end{align}
Here, $Contrast(\cdot)$ is designed to perform structural comparative analysis across domains, facilitating the discovery of domain-invariant patterns. $\mathcal{M}_{G_S}$ is expected to generalize to $\mathbb{D}_T$ despite domain-specific differences.

For instance, GCC~\cite{qiu2020gcc} leverages two structural augmentations to generate numerous subgraph instances from source domains. Positive pairs are formed from instances of the same graph, while negative pairs come from different graphs. These pairs are used to pretrain GNNs with a contrastive objective, enabling the model to capture structural commonalities in source domains and transfer them to target domains via fine-tuning or prompting. PCRec~\cite{wang2021pcrec} constructs positive and negative sample pairs by generating two augmented views of user-item graphs (without node features) in source domains, and uses contrastive learning to capture common structural patterns. It then transfers the pre-trained graph encoder to the target domain by fine-tuning it on target data, improving recommendation performance. GRADE~\cite{wu2023GRADE} introduces a Graph Subtree Discrepancy (GSD) metric, inspired by the Weisfeiler-Lehman test, to measure differences in structure between source and target graphs. Then, it trains a graph model by minimizing both the widely used task-specific loss on the source graph and this GSD term, which implicitly treats similar subtrees as positive pairs (pulled together) and dissimilar ones as negative pairs (pushed apart), enabling the trained model to capture the cross-domain structural commondalities across domains and be generalized to target domains effectively. FedStar~\cite{tan2023FedStar} leverages federated learning to capture shared structural patterns across domains. Specifically, each client locally trains a structural encoder to capture its domain-specific graph structures. Then, the server aggregates these encoders' parameters into a unified global model through weighted averaging. This iterative process implicitly reinforces structural patterns common to multiple domains (pulling "positive samples" closer) while suppressing domain-specific structural features (pushing "negative samples" apart), thereby achieving an implicit cross-domain structural contrastive transfer. APT~\cite{xu2023APT} proposes a graph selection strategy based on structural properties such as average degree and network entropy, aiming to select the most informative source graphs for contrastive pre-training. By constructing positive and negative sample pairs from these representative graphs, the model can be pre-trained more effectively to capture structural commonalities across domains and improve generalization to target graphs. BooG~\cite{cheng2024BooG} boosts the cross-domain learning of graph foundation models by unifying structural characteristics through virtual super nodes and edges, and employs a novel contrastive learning method to construct positive and negative sample pairs and pre-train a graph model, thereby capturing cross-domain structure commonalities that can be generalized to target domains. ProCom~\cite{wu2024procom} introduces a dual‐level contrastive pre‐training (i.e., node‐to‐context and context‐to‐context levels) method to learn shared community structures, thereby improving the generalizability of pre-trained GNN-based community detection models to various target domains. RiemannGFM~\cite{sun2025RiemannGFM} captures cross-domain knowledge by learning a universal structural vocabulary of graph substructures (e.g., trees and cycles), embedding them into Riemannian manifolds, and finally leveraging geometric contrastive learning for pre-training a graph model that can be easily transferred to target domains.

\subsection{Discussions} 

\label{sec: structure-oriented discussion}
Structure-oriented approaches have demonstrated superior performance on realizing CDGL, being able to handle various complex graph structural information. Moreover, they, specifically the generative approaches, also exhibit strong flexibility, as the unified node connection rules can generate meaningful structures for the graphs from unseen domains. 
Another advantage of these methods is their ability to explore the potential for uncovering commonalities across graph data from different domains, with structural information serving as the foundation of such data. 
For graph data, while structural information forms the foundation, it is the feature information that plays a crucial role in conveying substantive meaning. However, structure-oriented approaches mainly focus on capturing structural information but sometimes ignore the feature information. Taking structure generation approaches as an example, they need to consider the coherence between the generated structure and feature information to ensure effective communication across graph data from different domains.

\subsection{Significance and Future Implications}
Existing structure-oriented CDGL methods hold significant potential for bridging diverse domains by capturing universal structural insights. They effectively uncover structural commonalities across domains and achieve generalization at an open scale. Although relying solely on structural information may limit performance, these approaches provide critical insights into transferring structural patterns, paving the way for open CDGL and graph foundation models that seamlessly integrate structural and feature information.
\section{Feature-oriented Approach}
\label{sec:feature}

As discussed in Section~\ref{sec: structure-oriented discussion}, feature information plays a vital role in conveying substantive meaning in graph data, helping to differentiate between domains. The essence of feature-oriented approaches lies in learning transferable feature representations by identifying feature-level commonalities across domains. However, training a general graph model on data from multiple domains poses significant challenges, primarily due to differences in feature semantics and dimensions. 
To address these challenges, existing feature-oriented CDGL approaches can be categorized into two groups: \textbf{In-space} and \textbf{Cross-space}, depending on whether they require alignment of feature dimensions. In most cases, if the feature dimensions of source domain $A$ and target domain $B$ differ, their semantic meanings will inevitably differ as well, since it is highly unlikely that domains with different dimensions share the same semantics—this scenario typically requires \textbf{Cross-space} approaches. On the other hand, when $A$ and $B$ have the same feature dimensions, the semantics may either be identical or distinct. If the semantics are aligned, no additional adjustments are needed, and the domains can essentially be treated as one. However, when the semantics differ, further steps are required to bridge the semantic gap—this case calls for \textbf{In-space} approaches.

\subsection{In-space}
\begin{figure}[t]
  \centering
  \includegraphics[width=0.45\textwidth]{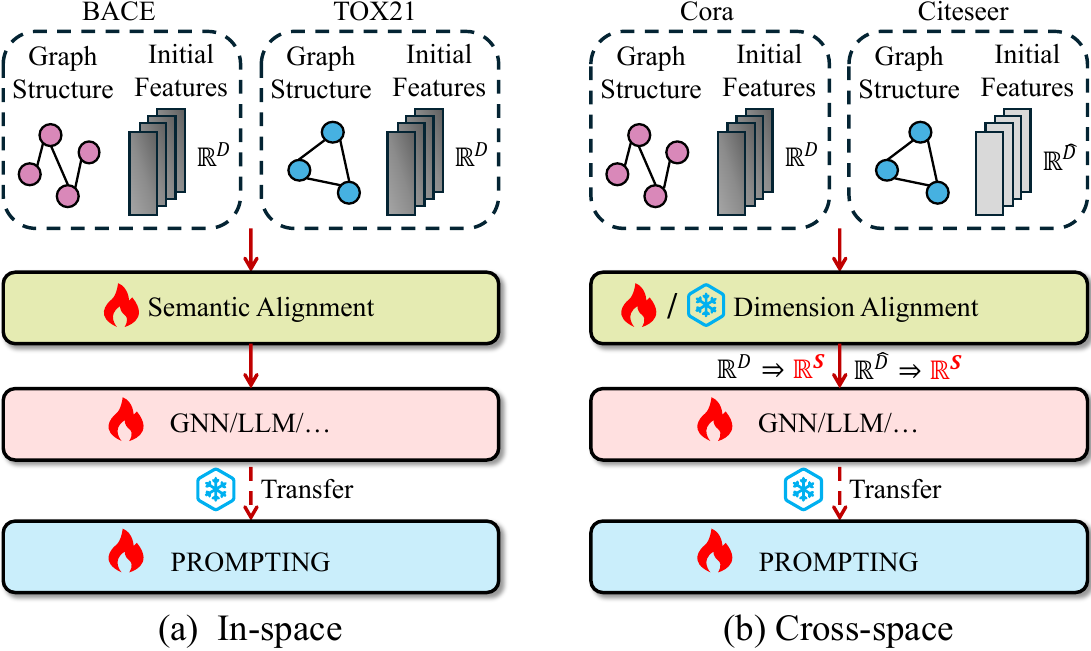}
  \caption{Feature-oriented CDGL: a) tackles the feature semantic alignment across domains; b) further aligns the feature dimensions across various graph domains to achieve more general CDGL. Prompt techniques are optional for both approaches.}
  \label{fig:overview of feature-oriented approaches}
\end{figure}

The earliest feature-oriented approaches are In-space, meaning they perform CDGL when the feature dimensions of the graph data in the source and target domains are consistent, primarily focusing on aligning the feature semantic information. As shown in Figure~\ref{fig:overview of feature-oriented approaches}, in-space methods typically begin by employing a semantic alignment module $SAlign(\cdot)$ to map the input graph feature $\mathcal{X}_S \in \mathbb{D}_{G_S}$ into a new feature $\mathcal{X}^{'}_S$, thereby resulting in new source domains $\mathbb{D}^{'}_{G_S}$. A graph model $\mathcal{M}_{\mathbb{D}^{'}_{G_S}}$ is then trained on the new source domains and subsequently generalized to the target domains $\mathbb{D}_{G_T}$ via a task-specific prediction head $TaskH(\cdot)$, which outputs the final prediction results. Besides, a promoting module $prompt(\cdot)$ is optionally leveraged to further enhance the cross-domain transferability:
\begin{align}
    \text{Semantic Alignment:} \quad & \mathcal{X}^{'}_S = SAlign(\mathcal{X}_S), \\
    \text{Generalization:} \quad & \hat{y}_T = TaskH(\mathcal{M}_{\mathbb{D}^{'}_{G_S}}(\mathcal{G}^{'}_T)),\\
    \text{(Prompt):} \quad & \mathcal{X}^{p}_T = prompt(\mathcal{X}_T),\\
    & \mathcal{X}^{p^{'}}_T = SAlign(\mathcal{X}^{p}_T),\\
    & \hat{y}_T = TaskH(\mathcal{M}_{\mathbb{D}^{'}_{G_S}}(\mathcal{G}^{p^{'}}_T))
\end{align}
Here, $SAlign(\cdot)$ is designed to align feature semantics across domains. $\mathcal{M}_{\mathbb{D}^{'}_{G_S}}$ is expected to generalize to $\mathbb{D}_T$ despite semantic-specific differences. $prompt(\cdot)$ is used to modify $\mathcal{X}_T$ into $\mathcal{X}^{p}_T$ to better align with the semantic space captured by $\mathcal{M}_{\mathbb{D}^{'}_{G_S}}$, thereby facilitating cross-domain transfer.

Note that the source and target domains typically share a high-level domain (e.g., molecular domain~\cite{li2022molecularsmall}) when the feature dimensions align. Thus, in-space approaches are generally considered part of the limited CDGL. For instance, KTN~\cite{yoon2022KTN} introduces a trainable knowledge transfer network that models the theoretical transformation between feature extractors of different node types in a heterogeneous graph. Specifically, it learns to map the target node embeddings into the source domain feature space using adjacency matrices and a shared learnable transformation matrix, thereby aligning their semantic representations and enabling effective cross-domain feature semantic transfer. NaP~\cite{wang2024NAP} proposes a contrastive learning framework that dynamically re-labels semantically similar cross-domain negative pairs as positive ones based on embedding similarity. By doing so, it pre-trains a domain-invariant GNN encoder that implicitly captures shared feature-level semantic commonalities across source domains and reduces the distribution discrepancy. The pre-trained encoder can be directly generalized to target domains, achieving cross-domain feature semantic transfer. GPF-plus~\cite{fang2024gpfplus} and RELIEF~\cite{zhu2024relief} are graph prompt-based methods that introduce learnable prompt features to augment graph representations. By adding prompt tokens to the original node features in target domains, they guide the GNN encoder to align input representations with a pretrained semantic space. These prompts implicitly capture semantic commonalities learned from source domains and enable effective cross-domain semantic transfer. DAGPrompT~\cite{chen2025DAGPrompT} proposes a GLoRA module, inspired by the success of LoRA in NLP~\cite{hu2022loraInNLP}, to achieve cross-domain semantic alignment through low-rank matrix updates for adapting the feature semantics of target domains to the pre-trained GNNs on source domains. GADT3~\cite{pirhayatifard2025crossdomainGADT3} leverages test-time training to achieve cross-domain adaptation, where a homophily-based self-supervised loss guides a target-specific encoder to align domain-invariant node patterns with a frozen source decoder, thereby transferring anomaly-discriminative knowledge without access to source data or target labels.

\subsection{Cross-space}
Since in-space approaches are significantly limited in cross-domain scale compared to open CDGL, Cross-space approaches have been proposed in recent years to tackle the feature dimension misalignment problem. Specifically, these methods can map cross-domain graph features into a shared embedding space with the same dimension, which enables the graph model to take various cross-domain graphs as input, thereby learning shared feature patterns across domains and relaxing the limitation of cross-domain scale. 
As shown in Figure~\ref{fig:overview of feature-oriented approaches}, cross-space methods typically begin by employing a dimension alignment module $DAlign(\cdot)$ to map the input graph features from different domains with different dimensions, denoted as $\mathcal{X}_{S_1}, \mathcal{X}_{S_2} \in \mathbb{D}_{G_S}$, $\mathcal{X}_{S_1} \in \mathbb{R}^{D}$ and $\mathcal{X}_{S_2} \in \mathbb{R}^{\hat{D}}$, into new ones $\mathcal{X}^{'}_{S_1}, \mathcal{X}^{'}_{S_2} \in \mathbb{R}^{S}$, thereby resulting in new source domains $\mathbb{D}^{'}_{G_S}$. A graph model $\mathcal{M}_{\mathbb{D}^{'}_{G_S}}$ is then trained on the new source domains and subsequently generalized to the target domain $\mathbb{D}_T$ via a task-specific prediction head $TaskH(\cdot)$, which outputs the final prediction results. Besides, a promoting module $prompt(\cdot)$ is also optional:
\begin{align}
    \text{Dimension Alignment:} \quad & \mathcal{X}^{'}_{S_1} = DAlign(\mathcal{X}_{S_1}), \\
    & \mathcal{X}^{'}_{S_2} = DAlign(\mathcal{X}_{S_2}), \\
    \text{Generalization:} \quad & \mathcal{X}^{'}_{T} = DAlign(\mathcal{X}_{T}), \\
    & \hat{y}_T = TaskH(\mathcal{M}_{\mathbb{D}^{'}_{G_S}}(\mathcal{G}^{'}_T)),\\
    \text{(Prompt):} \quad & \mathcal{X}^{p}_T = prompt(\mathcal{X}^{'}_T),\\
    & \hat{y}_T = TaskH(\mathcal{M}_{\mathbb{D}^{'}_{G_S}}(\mathcal{G}^{p}_T))
\end{align}
Here, $DAlign(\cdot)$ is designed to align feature dimensions across domains. $\mathcal{M}_{\mathbb{D}^{'}_{G_S}}$ is expected to generalize to $\mathbb{D}_T$ despite dimension-specific differences. $prompt(\cdot)$ is used to modify $\mathcal{X}_T$ into $\mathcal{X}^{p}_T$ to better align with the feature space captured by $\mathcal{M}_{\mathbb{D}^{'}_{G_S}}$.

For instance, OFA~\cite{liu2024OFA} and ZeroG~\cite{li2024zerog} both leverage LLMs to align text-attributed feature dimensions across domains. Particularly, OFA~\cite{liu2024OFA} freezes an LLM to generate LLM-based embeddings as node features for both source and target graphs, which are then fed into a lightweight classification head trained on source domains, enabling few-shot or zero-shot node classification on unseen target domains. Compared with OFA, ZeroG~\cite{li2024zerog} executes parameter-efficient fine-tuning (LoRA) on a frozen LLM, aligning node attribute and class description embeddings from source domains into a shared feature space. Then, it further enriches LLM-based node features by a semantic prompt node, and finally performs zero-shot node classification by measuring text similarity between node embeddings and class embeddings. Similar to the former approaches, GraphAlign~\cite{hou2024graphalign} also applies LLMs to align node feature dimensions, while additionally designing alignment strategies for feature encoding and normalization, alongside a mixture-of-feature-expert module for better handling cross-domain semantic information beyond dimension differences across domains to transfer feature commonalities. CDFS-GAD~\cite{chen2024CDFSGAD} employs a domain-adaptive graph contrastive learning module to pre-train a graph model, which can align the dimensions of graphs from various source domains and then construct positive and negative sample pairs across domains. Then, a domain-specific prompt-tuning module is introduced to align the features of target domains with pre-trained feature representation space while preserving their unique features. UNPrompt~\cite{niu2024UNPrompt} first applies feature projection (e.g., SVD~\cite{stewart1993SVD}) to align node feature dimensions across graph domains, then injects a small set of learnable prompt tokens into node attributes. These tokens are trained on a single source dataset and then frozen, enabling zero-shot anomaly detection by aligning the semantic space of target graphs with the pre-trained source representation.

\subsection{Discussions}
Compared with structure-oriented approaches, focusing on learning cross-domain feature commonalities to build graph models also demonstrates its own distinct advantages in handling CDGL, especially when the feature information between the source domains and target domains differs significantly. However, since feature-oriented approaches often skip the dedicated extraction of structural information, they may encounter conflicts between structural and feature patterns when training a cross-domain graph model, limiting the performance upper bound.

\subsection{Significance and Future Implications}
If structure-oriented CDGL research serves as the basis for achieving open CDGL or graph foundation models, then feature-oriented CDGL represents a critical breakthrough toward realizing true open cross-domain scalability. As mentioned earlier, the feature dimensions of graph data from different domains are often inconsistent, and even when they align, their semantic meanings may differ. This makes it exceedingly challenging to learn commonalities from cross-domain features. Existing feature-oriented CDGL works extensively explore how to comprehend feature information across domains and effectively capture their shared semantics. These insights are pivotal for advancing toward open CDGL, making feature-oriented approaches a key enabler of this vision.

\begin{figure*}[t]
  \centering
  \includegraphics[width=0.98\textwidth]{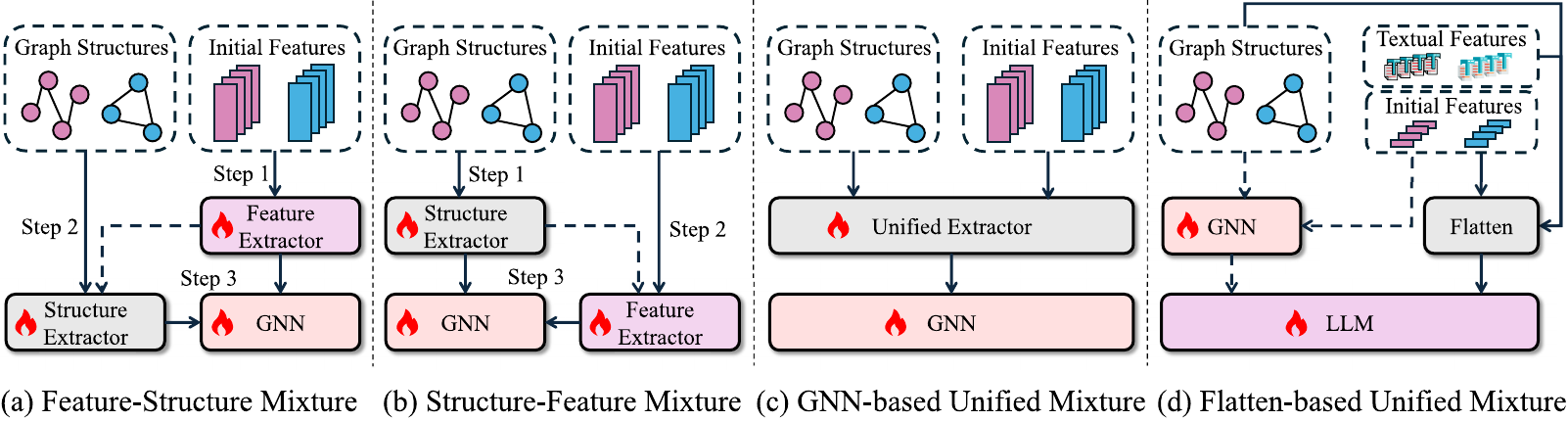}
  \caption{Mixture-oriented CDGL. (a) \textbf{Feature-Structure Mixture}: first extracts transferable features, then integrates structural knowledge. (b) \textbf{Structure-Feature Mixture}: first extracts shared structural patterns, then incorporates feature information. (c) \textbf{GNN-based Unified Mixture}: a unified extractor jointly learns cross-domain structural and feature patterns.  (d) \textbf{Flatten-based Unified Mixture}: graphs are flattened into token sequences, leveraging the powerful reasoning ability of LLMs to align and extract shared cross-domain knowledge. (a)-(b) follow a \emph{sequential-mixture} integration paradigm, whereas (c)-(d) adopt a \emph{unified-mixture} strategy. The red flame icon denotes trainable modules.} 
  \label{fig:overview of mixture-oriented approaches}
\end{figure*}
\section{Mixture-oriented Approach}
Mixture-oriented approaches are capable of performing cross-domain learning on both structural and feature information simultaneously, exploring the commonalities of structure and feature either individually or in pairs. This has been the most mainstream research direction in CDGL. Based on the different ways of fusing structure and feature information, we classify Mixture-oriented approaches into two types: Sequential Mixture and Unified Mixture.
\subsection{Sequential Mixture}
Similar to sequential ensemble learning techniques~\cite{rayana2016sequentialensemble1,yang2020sequentialensemble2}, the Sequential Mixture approach typically processes structural information and feature information in a sequential manner. 
As shown in Figure~\ref{fig:overview of mixture-oriented approaches}, sequential mixture approaches typically involve three steps: (1) employing a feature extractor $FExtract(\cdot)$ to extract shared graph features, mapping $\mathcal{X}_S \in \mathbb{D}_{G_S}$ into $\mathcal{X}^{'}_S$, (2) then, leveraging a structure extractor $SExtract(\cdot)$ to extract shared graph structures, mapping $\mathcal{A}_S \in \mathbb{D}_{G_S}$ into $\mathcal{A}^{'}_S$, and (3) finally, a graph model $\mathcal{M}_{\mathbb{D}^{'}_{G_S}}$ is trained on the new source domains $\mathbb{D}^{'}_{G_S}$ derived by $FExtract(\cdot)$ and $SExtract(\cdot)$ and subsequently generalized to the target domains $\mathbb{D}_T$ via a task-specific head function $TaskH(\cdot)$, which output the predicted results.
\begin{align}
    \text{Feature Extract:} \quad & \mathcal{X}^{'}_S = FExtract(\mathcal{X}_S), \\
    \text{Structure Extract:} \quad & \mathcal{A}^{'}_S = SExtract(\mathcal{A}_S), \\
    \text{Generalization:} \quad & \hat{y}_T = TaskH(\mathcal{M}_{\mathbb{D}^{'}_{G_S}}(\mathcal{G}^{'}_T,\mathcal{X}^{'}_T, \mathcal{A}^{'}_T)),
\end{align}
Here, the order of the feature extractor and the structure extractor can be swapped depending on the design.

For instance, SDGA~\cite{qiao2023SDGA} first leverages GCN to extract high-order structural information, followed by an adversarial transformation module with shift parameters to align the source graph feature semantic distribution with the target domain, thereby transferring structures and features in a sequential manner. 
GSPT~\cite{song2024GSPT} first samples node sequences via Node2Vec~\cite{grover2016node2vec} to capture cross-domain structural contexts, then maps node attributes into a unified space using a pretrained LLM (e.g., SentenceBERT~\cite{reimers2019sentenceBert}). It pre-trains a Transformer through masked feature reconstruction on these sequences to mine feature commonalities, and finally transfers to target domains without tuning, similar to ~\cite{liu2024OFA}.
SOGA~\cite{mao2024SOGA} aims to adapt a GNN pre-trained on source domains to target domains by leveraging the Structure Consistency and Information Maximization objectives. Specifically, the two objectives are computed separately for structures and features, thereby fine-tuning the pre-trained GNN to capture the structural and feature commonalities between the source and target domains, achieving cross-domain transfer. MLDGG~\cite{tian2025mldgg} sequentially learns domain-generalizable knowledge by first employing a structure learner to refine noisy edges and capture cross-domain structural patterns, and then applying a representation learner to disentangle domain-invariant semantics from domain-specific variations in node embeddings, achieving meta-level generalization to unseen domains.

Alternatively, the method can also prioritize the feature information before incorporating the structural information. An example of this is UniGraph~\cite{he2024unigraph}. Specifically, it first leverages a pretrained LM~\cite{he2020deberta,wang2022textE5} to unify node feature spaces on source domains. Then, it utilizes a GNN encoder to encode the structural information and introduces PPR-sampled subgraphs and a masked‐autoencoder learning objective to pretrain the LM and GNN encoder, distilling cross-domain commonalities. Finally, it can be effectively transferred to target domains via in-context learning or graph instruction tuning on an LLM~\cite{touvron2023llama, touvron2023llama2}.
PTGB~\cite{yang2023ptgb} first designs a data‐driven atlas mapping pipeline to project and align node features from multiple source domains into a common low-dimensional space, then pre-trains a GNN via a novel two-level contrastive learning objective to mine structural commonalities, and finally fine-tunes the pretrained GNN on target domains for cross-domain transfer.
ARC~\cite{liu2024ARC} first applies a smoothness-based feature alignment module to project and sort node attributes from source domains into a shared, anomaly-sensitive feature space; it then uses an ego-neighbor residual graph encoder to extract high-order structural anomaly patterns; finally, it performs anomaly detection on target domains via a cross-attentive in-context scoring module without any retraining.
AnyGraph~\cite{xia2024anygraph} first applies SVD to align the node-feature dimensions of multiple source domains into a shared embedding space. It then introduces a mixture-of-experts architecture—each specializing in a distinct structural pattern—and uses a lightweight routing mechanism to dynamically select the best expert for each graph. Finally, selected experts combined with corresponding feature encoders are jointly pre-trained on source domains to capture transferable cross-domain commonalities for target domains. 
GraphLoRA~\cite{yang2024graphlora} first injects a low-rank adaptation network alongside the frozen pre-trained GNN and then employs a structure-aware Maximum Mean Discrepancy (SMMD) loss to align node feature distributions between source and target graphs; it then uses structure-aware contrastive learning to infuse structural knowledge from the target graph into the model, ensuring effective cross-graph transfer at both structural and feature levels.
SAMGPT~\cite{yu2025SAMGPT} first aligns and adapts features across domains, then learns universal structure tokens during multi-domain pre-training, and finally utilizes holistic and specific prompts to transfer both unified and domain-specific structural knowledge to target domains.
MDGFM~\cite{wang2025MDGFM} first projects each source domain’s features into a shared tokenized semantic space to capture cross-domain semantic commonalities, then introduces the topology-aware contrastive learning strategy to align the graph structures on source domains, and finally transfers the learned tokens and GNN encoder to target domains via prompting.

\subsection{Unified Mixture}
In contrast to the Sequential Mixture approach, the Unified Mixture approach typically considers structural and feature information from a unified perspective during the training process, with the goal of achieving an organic mixture. As shown in Figure~\ref{fig:overview of mixture-oriented approaches}, unified mixture approaches typically involve two steps: (1) employing a unify function $Unify(\cdot)$ to extract graph structures and features simultaneously, mapping $\mathcal{A}_S \in \mathbb{D}_{G_S}$ and $\mathcal{X}_S \in \mathbb{D}_{G_S}$ into $\mathcal{A}^{'}_S$ and $\mathcal{X}^{'}_S$, and (2) then, a graph model $\mathcal{M}_{\mathbb{D}^{'}_{G_S}}$ is trained on the new source domains $\mathbb{D}^{'}_{G_S}$ derived by $Unify(\cdot)$ and subsequently generalized to the target domains $\mathbb{D}_T$ via a task-specific head function $TaskH(\cdot)$, which output the predicted results.
\begin{align}
    \text{Mixture Unifying:} \quad & \mathcal{X}^{'}_S,\mathcal{A}^{'}_S = Unify(\mathcal{X}_S,\mathcal{A}_S), \\
    \text{Prediction:} \quad & \hat{y}_T = TaskH(\mathcal{M}_{\mathbb{D}^{'}_{G_S}}(\mathcal{G}^{'}_T,\mathcal{X}^{'}_T, \mathcal{A}^{'}_T)),
\end{align}

Traditionally, the strategies for $Unify(\cdot)$ and $TaskH(\cdot)$ are part of a cohesive framework. The $Unify(\cdot)$ strategy is typically built upon the message-passing mechanism, where the differences lie in the ways of passing or the applied information encoding techniques. $TaskH(\cdot)$ is typically implemented using GNN-based heads (e.g., MLPs), which take the representations from $\mathcal{M}_{\mathbb{D}^{'}_{G_S}}$ and output predictions for the target domains $\mathbb{D}_T$. However, with the advent of the powerful representation and reasoning capabilities of LLMs in NLP, flatten-based unifying strategies have been introduced into graph learning. These strategies can fundamentally transform graph data into semantically rich node or token sequences. LLMs then leverage these sequences to perform reasoning and generate corresponding task results \cite{ijcai2024GFMSurvey}. To help comprehensively understand these unified mixture approaches, the following sections will illustrate these two sets of unified-task strategies in detail.

\paratitle{Graph-based Unified}

In general, GNN-based unified mixture approaches usually adopt similar selection strategies of task head functions while differing according to the message passing ways or the information encoding techniques. For example, DA-GCN~\cite{Guo2021DAGCN}, UDA-GCN~\cite{wu2020UDAGCN}, COMMANDER~\cite{ding2021COMMANDER}, CCDR~\cite{xie2022CCDR}, GAST~\cite{zhang2022GAST}, DH-GAT~\cite{xu2023DHGAT}, ALEX~\cite{yuan2023alex}, ACT~\cite{wang2023ACT}, ALCDR~\cite{zhao2023ALCDR}, DGASN~\cite{shen2023DGASN}, CMPGNN~\cite{wang2025bridgingCMPGNN}, and AEGOT-CDKE~\cite{wu2025crossAEGOTCDKT} all adopt domain adaptation training frameworks, differing mainly in their approaches to unifying data from source and target domains. Specifically, DA-GCN~\cite{Guo2021DAGCN} and UDA-GCN~\cite{wu2020UDAGCN} utilize graph convolutional networks (GCNs)~\cite{kipf2016gcn} with domain adaptation, which explores the initial possibilities for Unified Mixture CDGL by employing GCNs to aggregate structural and feature information together in a preliminary unified manner. To better handle the unified structure and feature information, more advanced methods have been introduced, such as graph attention mechanisms~\cite{velivckovic2018gat}, hyperbolic representations, and contrastive learning. COMMANDER~\cite{ding2021COMMANDER} employs a graph-attentive encoder for domain alignment, providing more flexibility and dynamic focus during feature aggregation. CCDR~\cite{xie2022CCDR} proposes a unified contrastive learning framework for cross-domain recommendation, utilizing both intra-domain and inter-domain contrastive learning to enhance representation learning and effectively transfer user preferences between domains. GAST~\cite{zhang2022GAST} uses a hybrid graph attention mechanism along with a POS-Transformer to integrate syntactic and sequential semantics, enhancing both structural and semantic representation. DH-GAT~\cite{xu2023DHGAT} applies hyperbolic graph attention networks to model hierarchical structures effectively, leveraging the power of hyperbolic space to represent latent relationships. ALEX~\cite{yuan2023alex} leverages SVD~\cite{stewart1993SVD} and domain discriminators to unify cross-domain representations, facilitating effective feature alignment. ACT~\cite{wang2023ACT} employs contrastive learning with domain alignment for anomaly detection, focusing on learning similarities and differences between domains to generate cohesive representations. ALCDR~\cite{zhao2023ALCDR} uses an optimal transport-based anchor link learning mechanism to adaptively transfer knowledge across domains, providing a robust means of cross-domain unification. DGASN~\cite{shen2023DGASN} employs a graph attention network~\cite{velivckovic2018gat} to simultaneously learn node and edge embeddings, and uses adversarial domain adaptation to align the source and target distributions, thus achieving domain-invariant graph representations. CMPGNN~\cite{wang2025bridgingCMPGNN} inserts cross-domain edges via a link predictor to enable message passing between domains, unifying structure and feature spaces while using a mutual information loss to preserve discriminativeness and mitigate domain distribution shifts. AEGOT-CDKT~\cite{wu2025crossAEGOTCDKT} achieves cross-domain transfer by aligning node embeddings from source and target domains using Graph Optimal Transport (GOT), reducing domain discrepancies, and enhancing knowledge tracing, modeled as node classification.

Additionally, CDGEncoder~\cite{hassani2022CDGEncoder}, PGPRec~\cite{yi2023PGPRec}, STGP~\cite{hu2024stgp}, GCOPE~\cite{zhao2024GCOPE}, UniGLM~\cite{fang2024uniglm}, CrossLink~\cite{huang2024CrossLink}, MDGPT~\cite{yu2024MDGPT}, MDP-GNN~\cite{lin2025MDPGNN}, UniGraph2~\cite{he2025unigraph2}, and GIT~\cite{wang2025GIT} all involve a pretraining phase, but they differ in their information encoding or propagation methods during pretraining. Specifically, CDGEncoder~\cite{hassani2022CDGEncoder} uses a multi-view GNN encoder combined with attention mechanisms to aggregate contextual and topological views. PGPRec~\cite{yi2023PGPRec} introduces personalized graph prompts and uses contrastive learning during pretraining to enhance user/item representations. STGP~\cite{hu2024stgp} utilizes a two-stage prompting strategy to adapt both domain and task-specific properties across spatio-temporal tasks. GCOPE~\cite{zhao2024GCOPE} employs graph coordinators to align diverse graph datasets during pretraining, mitigating negative transfer. UniGLM~\cite{fang2024uniglm} considers introducing LLMs as the encoder for text attributes in the traditional pre-training paradigm with a tailored contrastive learning approach based on the adaptive positive sample selection to align features and structures. CrossLink~\cite{huang2024CrossLink} leverages a decode-only transformer to model the temporal evolution of dynamic graphs for link prediction. Finally, MDGPT~\cite{yu2024MDGPT} uses domain tokens and dual prompts to unify and adapt multi-domain graph features, enhancing cross-domain generalization. MDP-GNN~\cite{lin2025MDPGNN}, a cross-domain pre-training framework, first learns a feature converter for each domain to map node attributes into a shared semantic space, then injects a set of shared pivot nodes whose learned links blend graph structures across domains, and finally minimizes a Wasserstein-distance term to align their overall distributions, where the pre-trained GNN model can be effectively fine-tuned or prompted in cross-domain downstream tasks.
UniGraph2~\cite{he2025unigraph2} focuses on cross-domain multimodal graph learning. It introduces a mixture-of-experts module~\cite{wang2023MoEs} to align and fuse cross-domain features, and trains a GNN with a unified self-supervised loss that reconstructs node attributes as well as graph structures, thus preserving rich semantic and structural information. The same pipeline also works for single-modality graphs by simply disabling the unused modality branch.
Compared with GFT~\cite{wang2024gft}, which performs cross-domain learning by directly converting each whole graph into a computation tree (a structure generation CDGL approach), GIT~\cite{wang2025GIT} first builds a task tree—adding a virtual task node and linking only task-relevant parts of the graph—then pre-trains a GNN on many such task trees from diverse domains with a contrastive reconstruction loss~\cite{zhu2020GRACEContrastive}, optionally instruction-tunes the encoder with a handful of labeled task trees from the target domain, and finally solves downstream tasks by transforming each new graph into its task tree and fine-tuning the pre-trained GNN model plus a lightweight classifier. OMOG~\cite{liu2024OMOG} first leverages a frozen LLM (i.e., Sentence-BERT)\cite{reimers2019sentenceBert} to unify node text features and applies multi-hop SGC\cite{wu2019SGC} to encode structural information from cross-domain graphs. It then pre-trains a dedicated expert model for each source graph to capture domain-specific knowledge. By computing the adaptability between these expert models and input graphs from new domains, OMOG enables effective few-shot and zero-shot learning on downstream tasks.

In addition to the above two mainstream paradigms, domain generalization, multi-task learning, and federated learning have also been explored. For example, CrossHG-Meta~\cite{zhang2022CrossHGMeta} proposes to process structural and feature information simultaneously by aggregating heterogeneous information from multiple semantic contexts and then designs a unified domain generalization framework for CDGL, especially for heterogeneous graphs. METABrainC~\cite{yang2022METABrainC} proposes a pre-training cross-domain framework via meta-learning techniques using multiple self-supervised tasks on source datasets and adapts it to target tasks via fine-tuning, enhancing cross-domain brain connectome analysis. GCFL~\cite{xie2021gcfl} employs a clustering algorithm to group graph models trained on clients based on parameter similarities, thereby achieving a unified representation of structural and feature information within a federated learning framework, ultimately enhancing cross-domain generalization.

\paratitle{Flatten-based Unified}

Referring to the input of LLMs, flatten-based Unified Mixture approaches focus on converting graphs into textual descriptions by transforming them into sequences of nodes or tokens. These approaches then develop tailored learning strategies to allow LLMs to process these sequences and generate task-specific outputs effectively.
For instance, GIMLET~\cite{zhao2023gimlet} introduces a unified graph-text approach called Generalized Position Embedding, which transforms graphs into token sequences and seamlessly combines graph structure with LLM-driven text processing.
GraphTranslator~\cite{zhang2024graphtranslator} employs a Translator module to convert node embeddings into token sequences, enabling effective integration of pre-trained graph models and LLMs, allowing LLMs to handle both pre-defined and open-ended graph tasks.
GraphGPT~\cite{tang2024graphgpt} leverages a dual-stage instruction tuning mechanism to align graph structural and semantic information with LLMs, enhancing cross-domain generalization through both graph structural alignment and node feature transfer.
GITA~\cite{wei2024gita} combines graph visualization with textual conversion, using a Vision-Language Model (VLM) to jointly reason over visual and textual graph data, providing a comprehensive integration of graph semantics and structure.
InstructGraph~\cite{wang2024instructgraph} uses a structured format verbalizer to convert graph data into text sequences, combined with instruction tuning and preference optimization to improve graph reasoning and reduce hallucinations, making LLMs suitable for graph-centric tasks.
TEA-GLM~\cite{wang2024TeaGLM} employs a linear projection module to align pre-trained Graph Neural Network (GNN) representations with LLM token embeddings, enabling zero-shot predictions across multiple graph tasks without modifying the LLM itself.
GOFA~\cite{kong2025gofa} introduces a hybrid architecture that interleaves GNN layers with a pre-trained LLM, combining structural and semantic modeling abilities while transforming graph data into flattened text descriptions for LLM-based text generation.
HiGPT~\cite{tang2024higpt} utilizes a heterogeneous graph tokenizer to transform complex graph structures into textual representations, employing a novel instruction-tuning framework to enhance LLM's understanding of intra- and inter-node relationships.
GraphCLIP~\cite{zhu2024graphclip} uses a graph-summary contrastive learning approach by generating textual descriptions of subgraphs with an LLM, which improves zero-shot and few-shot transfer capabilities through graph prompt tuning and invariant learning.
GraphWiz~\cite{chen2024graphwiz} leverages the GraphInstruct dataset to train LLMs in solving graph computational problems, using an instruction-following framework that transforms graphs into textual sequences, allowing explicit reasoning paths for improved task performance.

\vspace{-1.25em}
\subsection{Discussions}

Mixture-oriented approaches mainly focus on simultaneously integrating structure and feature information across domains, enabling deep learning models to more comprehensively capture commonalities across domains. Nevertheless, the absence of a powerful feature alignment technique that effectively maps datasets from different domains into a shared feature space remains a significant limitation for feature-oriented approaches. As shown in Table~\ref{tab:summarizations}, earlier mixture-oriented approaches were only able to scale to limited cross-domain settings. Recently, some feature alignment techniques, such as large language models (LLMs), have been introduced, but they have only managed to achieve performance at a conditional scale. Indeed, a few newly proposed methods have demonstrated the potential to achieve open cross-domain scale, but their primary focus has been on exploring the feasibility of true open CDGL. For example, some approaches utilize domain-specific linear projection layers to align feature dimensions across domains, yet the resulting performance improvements have been limited.
\vspace{-1.5em}
\subsection{Significance and Future Implications}
Mixture-oriented approaches are fundamentally designed to effectively integrate structure-oriented and feature-oriented methods, aiming to achieve a synergistic effect where the whole is greater than the sum of its parts (i.e., $1+1>2$). For instance, since the cross-domain representation spaces derived from structural and feature information are inherently distinct, finding an organic way for these spaces to adapt to each other is both challenging and crucial. Current explorations in mixture-oriented approaches not only offer valuable insights but also pave the way for achieving truly open CDGL and even the development of graph foundation models.

\section{Open Questions and Insights}

Each of the three paradigms reflects a distinct philosophy on how transferable knowledge can be shared and generalized across heterogeneous graph domains. Building upon the systematic review of existing works, we further summarize shared insights and emerging challenges. Specifically, this section discusses three fundamental questions central to CDGL research: 
\textbf{(1) What types of graph structural information (e.g., topology regularities) and feature representations are transferable across domains?} This concerns identifying the domain-invariant components that generalize beyond domain-specific graph construction rules. 
\textbf{(2) What is the nature of the knowledge being transferred across domains?} Beyond explicit structures or features, transferable knowledge may manifest as higher-level statistical priors, semantic abstractions, or topological invariances that implicitly support domain generalization. 
\textbf{(3) Can we truly achieve open cross-domain transfer across graphs that are not originally designed in graph form?} This raises a critical question about whether a general-purpose graph foundation model can effectively bridge domains constructed from fundamentally different modalities, such as those derived from NLP or CV.

Recent advances across the three paradigms collectively indicate that both structural and feature information contain domain-invariant elements for cross-domain transfer. From the structural perspective, shared connectivity patterns such as local motifs, hierarchical topologies, or community structures tend to preserve universal relational principles, making structural transfer inherently easier even without domain-specific semantics. For example, when ignoring domain features, two graphs can be directly compared through kernel-based similarity measures~\cite{nikolentzos2021kernelForSharedStructure1}, which standardize heterogeneous structures into comparable forms. This structural normalization allows models pretrained on source domains to more readily identify shared topological patterns in target domains. 

While structural transfer captures domain-invariant topology, the unified transfer of structural and feature information is far more critical in real-world scenarios, as features assign concrete semantic meanings to different topological structures. This enables models to derive more fine-grained node or graph representations that are meaningful across diverse domains. However, achieving such unified transfer is highly non-trivial, which explains why numerous studies focus on learning how to align structural and feature spaces effectively~\cite{yuanmuchUnifiedImportance1}. In particular, textual or tokenized features become more transferable when combined with LLMs, as they naturally reside within a shared embedding space where semantic correlations can guide structural reasoning~\cite{qi2025bridgeglmLanguageTransfer1}. Through such representations, models can learn how specific textual semantics correspond to particular topological patterns and how structural configurations evolve as feature meanings shift. In essence, the transferable knowledge in CDGL arises from the latent correspondence between topology and semantics, reflecting how semantic variations reshape structural patterns. Achieving robust CDGL thus requires models capable of discovering the underlying design principles of graphs and harmonizing structural and semantic signals into invariant, domain-agnostic representations.

Beyond the above two questions, a more fundamental challenge lies in determining whether true \textit{open} cross-domain transfer is achievable across graph data that are not originally designed in graph form. Existing open CDGL methods~\cite{zhao2024GCOPE} have primarily investigated transferring knowledge among various graph domains (e.g., homogeneous and heterogeneous graphs) that inherently follow graph construction principles. However, it remains unclear whether graphs derived from other modalities—such as those constructed from pure NLP corpora or CV datasets—can serve as compatible domains for transfer. Nevertheless, this challenge also highlights a potential opportunity: even for such secondarily designed graphs, the underlying design principles and the alignment between structural and semantic signals may still be transferable. For instance, text-attributed graphs are originally constructed from natural language data but have demonstrated transferability across textual domains~\cite{liu2024OFA,li2024zerog}. Similarly, visual graphs derived from image descriptions or region relationships can potentially leverage shared semantic representations through language grounding~\cite{wang2025multimodalgraphCVCDGL1}. These insights suggest that the boundary of graph domains may be extendable through cross-modal representations, offering a promising direction toward realizing genuine graph foundation models.

\begin{table*}[htbp]
\renewcommand{\arraystretch}{1.6}
\resizebox{0.98\textwidth}{!}{
\begin{tabular}{llllllllllllc}
\toprule
& \textbf{Model} & \textbf{GNN Backbone} & \textbf{Paradigm} & \textbf{Scale} & \textbf{Difficulty} & \textbf{LLM} & \textbf{Fine-tuning} & \textbf{Prompting} & \textbf{Domain} & \textbf{Setting} & \textbf{Task} & \textbf{Code} \\ \midrule
\multirow{10}{*}{\rotatebox{90}{Structure-oriented}} 
& GCC\cite{qiu2020gcc} & GIN & Pre-training & Open & High & - & \Checkmark & \XSolidBrush & Social/Academic Networks, etc. & Full-shot & Node, Graph & \href{https://github.com/THUDM/GCC}{Link} \\
& PCRec\cite{wang2021pcrec} & GIN & Pre-training & Limited & High & - & \Checkmark & \XSolidBrush & Recommendation Networks & Full-shot & Link & - \\
& APT\cite{xu2023APT} & GIN & Pre-training & Open & High & - & \Checkmark & \XSolidBrush & Citation/Protein Networks, etc. & Full-shot & Node, Graph & \href{https://github.com/galina0217/APT}{Link} \\
& GRADE\cite{wu2023GRADE} & GCN & Domain Adaptation & Open & Low & - & \XSolidBrush & \XSolidBrush & Citation /Recommendation Networks, etc. & Unsupervised & Node, Link & \href{https://github.com/jwu4sml/GRADE}{Link} \\
& FedStar\cite{tan2023FedStar} & GCN, GIN & Federated Learning & Open & Low & - & \Checkmark & \XSolidBrush & Biological/Social Networks, etc. & Full-shot & Graph & \href{https://github.com/yuetan031/FedStar}{Link} \\
& ProCom\cite{wu2024procom} & GCN & Pre-training & Conditional & High & - & \XSolidBrush & \Checkmark & Social/E-commerce, etc. & Few-shot & Community Detection & \href{https://github.com/WxxShirley/KDD2024ProCom}{Link} \\
& GraphControl\cite{zhu2024graphcontrol} & GIN & Pre-training & Open & High& - & \Checkmark & \Checkmark & Social/Networks, etc. & Full/Few-shot & Node, Graph & \href{https://github.com/wykk00/GraphControl}{Link} \\
& GA$^{2}$E\cite{hu2024GA2E} & GAE & Pre-training & Open & Moderate & - & \Checkmark & \Checkmark & Citation/Community Networks, etc. & Full-shot & Node, Link, Graph & - \\
& BooG\cite{cheng2024BooG} & GAT, GIN & Pre-training & Conditional & High & Sentence Transformer & \Checkmark & \Checkmark & Citation/Molecular, etc. & Zero/Few-shot & Node, Link, Graph & \href{https://anonymous.4open.science/r/BooG-EE42/}{Link} \\
& UniAug\cite{tang2024UniAug} & GCN, GIN & Pre-training & Open & Moderate & - & \XSolidBrush & \XSolidBrush & Biological/Chemical Networks, etc. & Full-shot & Node, Link, Graph & \href{https://github.com/WenzhuoTang/UniAug}{Link} \\
& GFT\cite{wang2024gft} & GraphSAGE & Pre-training & Open & High & - & \XSolidBrush & \XSolidBrush & Social/Molecular Networks, etc. & Few-shot & Node, Link, Graph & \href{https://github.com/Zehong-Wang/GFT}{Link} \\
& OpenGraph\cite{xia2024opengraph} & GT & Pre-training & Open & High & - & \XSolidBrush & \Checkmark & Recommendation/Social Networks, etc. & Zero/Few-shot & Node, Link & \href{https://github.com/HKUDS/OpenGraph}{Link} \\
& RiemannGFM\cite{sun2025RiemannGFM} & - & Pre-training & Open & High & - & \Checkmark & \XSolidBrush & Citation/Social Networks, etc. & Full/Few-shot & Node, Link & \href{https://github.com/RiemannGraph/RiemannGFM}{Link} \\ 
& GFSE\cite{chen2025gfse} & GT & Pre-training & Open & High & - & \Checkmark & \XSolidBrush & Citation/Molecular Networks, etc. & Few-shot & Node, Graph & \href{https://github.com/Graph-and-Geometric-Learning/GFSE}{Link} \\ \midrule
\multirow{7}{*}{\rotatebox{90}{Feature-oriented}} 
& KTN\cite{yoon2022KTN} & HGNN & Domain Adaptation & Limited & Low & - & \XSolidBrush & \XSolidBrush & Academic Networks & Zero-shot & Node & - \\
& NaP\cite{wang2024NAP} & GCN & Domain Generalization & Limited & High & - & \XSolidBrush & \XSolidBrush & Social Networks & Full-shot & Node & - \\
& CDFS-GAD\cite{chen2024CDFSGAD} & GraphSAGE & Domain Adaptation & Open & Low & - & \XSolidBrush & \Checkmark & Social Networks & Few-shot & Graph Anomaly Detection & - \\
& OFA\cite{liu2024OFA} & GCN, GAT, etc. & Pre-training & Conditional & High & Sentence Transformer & \XSolidBrush & \Checkmark & Citation/Molecular Networks, etc. & Zero/Few-shot & Node, Link, Graph & \href{https://github.com/LechengKong/OneForAll}{link} \\
& ZeroG\cite{li2024zerog} & GCN, GAT, etc. & Pre-training & Conditional & High & Sentence Transformer & \XSolidBrush & \Checkmark & Citation Networks & Zero-shot & Node & \href{https://github.com/NineAbyss/ZeroG}{link} \\
& GraphAlign\cite{hou2024graphalign} & GCN & Pre-training & Open & High & E5-small & \XSolidBrush & \XSolidBrush & Citation Networks, Knowledge Graphs & Few-shot & Node, Link & \href{https://github.com/THUDM/GraphAlign}{link} \\
& RELIEF\cite{zhu2024relief} & GCN & Pre-training & Limited & High & - & \XSolidBrush & \Checkmark & Citation Networks, Molecular, etc. & Few-shot & Node, Graph & \href{https://github.com/JasonZhujp/RELIEF}{link} \\
& GPF-plus\cite{fang2024gpfplus} & GIN & Pre-training & Limited & Moderate & - & \Checkmark & \Checkmark & Citation Networks, Knowledge Graphs, etc. & Few-shot & Node, Link & \href{https://github.com/zjunet/GPF}{link} \\
& DAGPrompT\cite{chen2025DAGPrompT} & GCN & PEFT & Limited & High & - & \Checkmark & \Checkmark & Citation/Molecular Networks, etc. & Full/Few-shot & Node,Graph & \href{https://github.com/Cqkkkkkk/DAGPrompT}{Link} \\ 
& UNPrompt\cite{niu2024UNPrompt} & GNN & Pre-training & Limited & High & - & \Checkmark & \Checkmark & Social Networks, etc. & Zero-shot & Node & \href{https://github.com/mala-lab/UNPrompt}{Link} \\
& GADT3\cite{pirhayatifard2025crossdomainGADT3} & GraphSage & Domain Adaptation & Open & High & - & \XSolidBrush & \XSolidBrush & Social Networks, etc. & Full-shot & Graph Anomaly Detection & \href{https://github.com/delaramphf/GADT3-Algorithm}{Link} \\ \midrule
\multirow{38}{*}{\rotatebox{90}{Mixture-oriented}} 
& UDA-GCN\cite{wu2020UDAGCN} & GCN & Domain Adaptation & Limited & Low & - & \XSolidBrush & \XSolidBrush & Citation Networks & Full-shot & Node & - \\
& COMMANDER\cite{ding2021COMMANDER} & GAT & Domain Adaptation & Limited & Low & - & \XSolidBrush & - & Social Networks & Unsupervised & Graph Anomaly Detection & - \\
& DA-GCN\cite{Guo2021DAGCN} & GCN & Domain Adaptation & Limited & Low & - & \XSolidBrush & \XSolidBrush & Recommendation Networks & Full-shot & Node & - \\
& GCFL\cite{xie2021gcfl} & GIN & Federated Learning & Open & Low & - & \XSolidBrush & \XSolidBrush & Molecular/Social Networks & Full-shot & Graph & - \\
& METABrainC\cite{yang2022METABrainC} & GCN, GAT, etc. & Multi-task learning & Limited & High & - & \Checkmark & \XSolidBrush & Brain/Molecular Networks, etc. & Few-shot & Graph\ & - \\
& GAST\cite{zhang2022GAST} & HGAT & Domain Adaptation & Limited & Low & GloVe, BERT & \Checkmark & \XSolidBrush & Review Networks & Full-shot & Sentiment Classification & - \\
& CDGEncoder\cite{hassani2022CDGEncoder} & GIN & Domain Adaptation & Open & Low & - & \Checkmark & \XSolidBrush & Molecular/Biological Networks, etc. & Few-shot & Graph & - \\
& CrossHG-Meta\cite{zhang2022CrossHGMeta} & HGNN & Domain Generalization & Limited & High & - & \XSolidBrush & \XSolidBrush & Academic Networks & Few-shot & Node & \href{https://github.com/aslandery/CrossHG-Meta}{Link} \\
& CCDR\cite{xie2022CCDR} & GAT & Contrastive Learning & Conditional & High & - & \XSolidBrush & \XSolidBrush & Recommendation Networks & Zero-shot & Link & \href{https://github.com/lqfarmer/CCDR}{Link} \\
& PTGB\cite{yang2023ptgb} & GCN, GAT, etc. & Pre-training & Limited & High & - & \Checkmark & \XSolidBrush & Brain Networks & Full-shot & Node, Graph & \href{https://github.com/Owen-Yang-18/BrainNN-PreTrain}{Link} \\
& SDGA\cite{qiao2023SDGA} & GCN & Domain Adaptation & Limited & Low & - & \XSolidBrush & \XSolidBrush & Citation Networks & Semi-supervised & Node & - \\
& ALCDR\cite{zhao2023ALCDR} & LightGCN & Domain Adaptation & Conditional & Low & - & \XSolidBrush & \XSolidBrush & Recommendation Networks & Full-shot & Link & - \\
& DGASN\cite{shen2023DGASN} & GAT & Domain Adaptation & Limited & Low & - & \XSolidBrush & \XSolidBrush & Citation Networks & Full-shot & Link & \href{https://github.com/Qqqq-shao/DGASN}{Link} \\
& CDTC\cite{zhang2023CDTC} & GIN & Domain Adaptation & Limited & Low & - & \XSolidBrush & \Checkmark & Molecular Networks & Few-shot & Graph & - \\
& ACT\cite{wang2023ACT} & GraphSAGE & Domain Adaptation & Limited & Low & - & \XSolidBrush & \XSolidBrush & Social Networks & Full/Few-shot & Graph & \href{https://github.com/QZ-WANG/ACT}{Link} \\
& DH-GAT\cite{xu2023DHGAT} & GAT & Domain Adaptation & Limited & Low & BERT & \Checkmark & \XSolidBrush & News/Social Media Graphs & Full-shot & Entity Recognition & - \\
& ALEX\cite{yuan2023alex} & GCN & Domain Adaptation & Conditional & Low & - & \XSolidBrush & \XSolidBrush & Citation/Academic Networks & Full-shot & Node & - \\
& PGPRec\cite{yi2023PGPRec} & GCN, GAT, etc. & Pre-training & Conditional & High & - & \XSolidBrush & \Checkmark & Recommendation Networks & Full-shot & Link & - \\
& CrossLink\cite{huang2024CrossLink} & DyGFomer & Domain Generalization & Open & High & - & \XSolidBrush & \XSolidBrush & Social Networks & Full-shot & Link & - \\
& GCOPE\cite{zhao2024GCOPE} & GCN, FAGCN & Pre-training & Open & High & - & \Checkmark & \Checkmark & Citation/Social Networks & Few-shot & Node, Graph & \href{https://github.com/cshhzhao/GCOPE}{Link} \\
& Uni-GLM\cite{fang2024uniglm} & GCN & Pre-training & Conditional & Moderate & BERT & \Checkmark & \XSolidBrush & Citation/E-commerce Networks & Semi-supervised, Few-shot & Node, Link & \href{https://github.com/NYUSHCS/UniGLM}{Link} \\
& SOGA\cite{mao2024SOGA} & GCN & Domain Adaptation & Limited & Low & - & \XSolidBrush & \XSolidBrush & Citation Networks & Unsupervised & Node & - \\
& STGP\cite{hu2024stgp} & Graph Transformer & Pre-training & Open & High & - & \XSolidBrush & \Checkmark &  Traffic Networks & Few-shot & Node & \href{https://github.com/hjf1997/STGP}{link} \\
& GSPT\cite{song2024GSPT} & - & Pre-training & Conditional & Moderate & SentenceBERT & \Checkmark & \Checkmark & Citation Networks & Few-shot & Node, Link & \href{https://github.com/SongYYYY/GSPT}{Link} \\
& ARC\cite{liu2024ARC} & GCN & Pre-training & Conditional & High & - & \XSolidBrush & \Checkmark & Citation/Social Networks, etc. & Few-shot & Graph Anomaly Detection & \href{https://github.com/yixinliu233/ARC}{link} \\
& UniGraph\cite{he2024unigraph} & GAT & Pre-training & Conditional & High & Llama, vicuna-7B, etc. & \Checkmark & \Checkmark & Citation/Molecular Networks, etc. & Zero/Few-shot & Node, Edge, Graph & \href{https://github.com/yf-he/UniGraph}{Link} \\
& TEA-GLM\cite{wang2024TeaGLM} & GraphSAGE & Pre-training & Conditional & High & Vicuna-7B & \XSolidBrush & \Checkmark & Citation/E-commerce Networks & Zero-shot & Node, Link & \href{https://github.com/W-rudder/TEA-GLM}{link} \\
& GOFA\cite{kong2025gofa} & TransConv & Pre-training & Conditional & High & Llama2, Mistral, etc. & \Checkmark & \Checkmark & Citation/Recommendation Networks & Zero-shot & Node, Link & \href{https://github.com/JiaruiFeng/GOFA}{Link} \\
& HiGPT\cite{tang2024higpt} & HGNN & Pre-training & Conditional & High & Sentence-BERT & \Checkmark & \Checkmark & Academic Networks & Few-shot/Zero-shot & Node & \href{https://github.com/HKUDS/HiGPT}{Link} \\
& MDGPT\cite{yu2024MDGPT} & GIN & Pre-training & Open & High & - & \XSolidBrush & \Checkmark & Citation/E-commerce Networks, etc. & Few-shot & Node, Graph & - \\
& AnyGraph\cite{xia2024anygraph} & GCN & Pre-training & Open & High & - & \XSolidBrush & \XSolidBrush & E-commerce/Academic Networks, etc. & Zero-shot & Node, Link, Graph & \href{https://github.com/HKUDS/AnyGraph}{Link} \\
& GIMLET\cite{zhao2023gimlet} & - & Pre-training & Limited & High & T5 & \Checkmark & \Checkmark & Molecular Networks & Few-shot/Zero-shot & Graph & \href{https://github.com/zhao-ht/GIMLET}{Link} \\
& GraphTranslator\cite{zhang2024graphtranslator} & GraphSAGE & Pre-training & Conditional & High & ChatGLM2-6B & \Checkmark & \Checkmark & E-commerce/Academic Networks & Few-shot/Zero-shot & Node & \href{https://github.com/alibaba/GraphTranslator}{Link} \\
& GraphGPT\cite{tang2024graphgpt} & Graph Transformer & Instruction tuning & Conditional & High & Vicuna-7B-v1.1/1.5 & \Checkmark & \Checkmark & Citation Networks & Zero-shot & Node, Link & \href{https://github.com/varunshenoy/GraphGPT}{Link} \\
& GITA\cite{wei2024gita} & GCN, GAT & Instruction tuning & Open & High & GPT-4V, LLaVA-7B/13B & \Checkmark & \Checkmark & Pure Structure Graphs & Zero-shot/Full-shot & Node, Link & \href{https://github.com/WEIYanbin1999/GITA/}{Link} \\
& InstructGraph\cite{wang2024instructgraph} & - & Instruction tuning & Conditional & High & LLaMA2 & \Checkmark & \Checkmark & Pure Structure Graphs, Citation Networks, etc. & Full/Few-shot & Node, Generation, Reasoning, etc. & \href{https://github.com/wjn1996/InstructGraph}{Link} \\
& GraphCLIP\cite{zhu2024graphclip} & GCN, GAT & Pre-traning & Conditional & High & Qwen2-72B & \Checkmark & \Checkmark & Academic/E-commerce Networks,etc & Few/Zero-shot & Node, Link, Graph & \href{https://github.com/ZhuYun97/GraphCLIP}{Link} \\
& GraphWiz\cite{chen2024graphwiz} & - & Instruction tuning & Limited & High & LLaMA2-7B/13B, Mistral-7B & \Checkmark & \Checkmark & Pure Structure Graphs & Zero-shot/Few-shot & Reasoning & \href{https://github.com/ZhuYun97/GraphCLIP}{Link} \\
& GraphLoRA\cite{yang2024graphlora} & GAT & PEFT & Open & High & - & \Checkmark & \XSolidBrush & E-commerce/Social Networks, etc. & Full/Few-shot & Node & \href{https://github.com/AllminerLab/GraphLoRA}{Link} \\
& OMOG\cite{liu2024OMOG} & SGC & Pre-training & Conditional & High & Sentence-BERT & \Checkmark & \XSolidBrush & Citation/Social Networks, etc. & Few/Zero-shot & Node, Link & - \\
& MDP-GNN\cite{lin2025MDPGNN} & GT & Pre-training & Open & Low & - & \Checkmark & \Checkmark & Citation/Social Networks, etc. & Few-shot & Node & - \\
& MDGFM\cite{wang2025MDGFM} & GCN & Pre-training & Open & High & - & \XSolidBrush & \Checkmark & Citation/Social Networks, etc. & Few-shot & Node & - \\
& SAMGPT\cite{yu2025SAMGPT} & GCN & Pre-training & Open & High & - & \XSolidBrush & \Checkmark & Citation/Social Networks, etc. & Few-shot & Node,Graph & \href{https://github.com/blue-soda/SAMGPT}{Link} \\ 
& UniGraph2\cite{he2025unigraph2} & GAT & Pre-training & Open & High & - & \Checkmark & \XSolidBrush & Citation Networks, Knowledge Graphs, etc. & Full/few-shot & Node,Graph & \href{https://github.com/yf-he/UniGraph2}{Link} \\
& GIT\cite{wang2025GIT} & GraphSAGE & Pre-training & Conditional & Hard & - & \Checkmark & \XSolidBrush & Citation/Molecular Networks, etc. & Full/few-shot & Node, Link, Graph & \href{https://github.com/Zehong-Wang/GIT}{Link} \\
& MLDGG\cite{tian2025mldgg} & GCN & Pre-training & Open & High & - & \Checkmark & \XSolidBrush & Social Networks & Full-shot & Node & - \\
& CMPGNN\cite{wang2025bridgingCMPGNN} & GCN, GIN & Domain Adaptation & Limited & High & - & \XSolidBrush & \XSolidBrush & Citation/Social Networks & Semi-supervised & Node & - \\
& AEGOT-CDKT\cite{wu2025crossAEGOTCDKT} & GAT & Domain Adaptation & Limited & Low & - & \Checkmark & \XSolidBrush & Educational Networks & Supervised & Node & - \\
\bottomrule
\end{tabular}}
\caption{A summary of models that leverage cross-domain knowledge to enhance graph-related tasks in the literature, ordered by their release time. \textbf{Paradigm} denotes the core technique used to achieve Cross-Domain Graph Learning. \textbf{Domain} indicates the domains across which knowledge is transferred. \textbf{Scale} and \textbf{Difficulty} refer to the cross-domain scales and the level of difficulty. \textbf{Fine-tuning} denotes whether it is necessary to fine-tune the parameters of Graph Neural Networks (GNNs). \textbf{Prompting} indicates the use of graph- or text-formatted prompts. Acronyms in \textbf{Task}: Node, Link, and Graph refer to node-level, link-level, and graph-level tasks, respectively. Beyond these general graph learning tasks, reasoning represents a higher-level reasoning ability on graphs, whereas other values correspond to specific application-oriented settings (e.g., graph anomaly detection).}
\label{tab:summarizations}
\end{table*}
\vspace{-0.5em}
\section{Future Directions}

\tabref{tab:summarizations} summarizes the graph models that transfer particular graph information to achieve CDGL according to the proposed taxonomy. Based on the review and analysis illustrated above, we believe that there is still much space for further enhancement in this field, especially for realizing the true graph foundation models. In this section, we discuss the remaining limitations of capturing graph information across domains used for achieving effective CDGL and list some directions for further exploration in subsequent research.
\vspace{-1em}

\subsection{Feature Alignment.} The correlation between feature semantics and dimensions across graph data from different domains is a crucial aspect to consider in CDGL. Misaligned feature dimensions will hinder the training of a general graph model capable of handling graph data from various domains, while misaligned semantics will prevent the trained graph model from capturing shared characteristics across domains, thereby limiting the performance of CDGL.

In existing CDGL methods, those targeting limited-scale scenarios can often skip this issue, as the graph data from source and target domains typically share identical feature dimensions by nature, such as in molecular networks~\cite{jin2024crossModalsurvey1,fang2024gpfplus}. For conditional scenarios, which mainly focus on graph data in recommendation systems~\cite{yi2023PGPRec} and TAGs~\cite{liu2024OFA,li2024zerog}, feature alignment challenges are circumvented in different ways: Graph data in recommendation systems may come from different domains, but they are all built on the basis of users and items, thus avoiding feature alignment problems. For TAGs, features can be easily aligned using embeddings generated by LLMs.

However, these methods are insufficient for addressing open CDGL, where feature dimensions and semantics are highly complex and heterogeneous across domains. Although some methods have begun exploring feature alignment solutions for more open scenarios, the research is still in its early stages. For instance, methods like ProG~\cite{sun2023prog}, MDGPT~\cite{yu2024MDGPT}, and CDFS-GAD~\cite{chen2024CDFSGAD} have attempted to align feature dimensions across different datasets using techniques such as SVD or MLP. However, even after aligning feature dimensions, the aligned features may still lack semantic equivalence. Therefore, proposing a universal feature alignment approach to handle datasets from different domains is a key research direction towards achieving effective open CDGL.
\vspace{-1em}
\subsection{Dealing with the Scale of Cross-domain Datasets.} In NLP and CV, the cross-domain generalization ability of foundation models follows the Scaling Law, which suggests that generalization improves with an increase in model size and the number of training samples. Typically, training an effective foundation model requires a large volume of samples, often in the order of billions, whereas traditional CDGL models generally rely on much smaller training datasets. This discrepancy in data volume limits the cross-domain performance and generalization capabilities of these models. While structure-oriented methods can process large-scale graph-structured data and achieve open CDGL at the structural level, the lack of feature information restricts their performance. However, when both structural and feature information are considered, traditional CDGL methods usually achieve only limited or conditional cross-domain learning, inherently constraining the diversity and scale of the datasets used for training. Although certain models, such as GCOPE, can alleviate some limitations on dataset diversity and scale to achieve open CDGL, the scale and diversity of the datasets remain limited. Therefore, to significantly improve the performance of graph models in open cross-domain scenarios and contribute to true graph foundation models, large-scale and diverse datasets are essential.
\vspace{-1em}
\subsection{Overcoming data leakage.} \tabref{tab:summarizations} demonstrates that LLMs are receiving increasing attention in CDGL, exploring their potential capabilities in this area. However, the introduction of LLMs inevitably brings about the issue of data leakage, which has become a focal point of discussion \cite{aiyappa2023LLMsImportant}. Specifically, since LLMs are pre-trained on extensive text corpora, it's likely that they may have seen and memorized at least part of the test data from common benchmark datasets, especially in citation networks. This undermines the reliability of current studies that rely on earlier benchmark datasets. Moreover, \cite{chen2024LLMsMemoryGraph} demonstrates that specific prompts could potentially enhance the "activation" of LLMs' corresponding memory, thereby influencing evaluation outcomes. Both \cite{huang2023dataleakage1} and \cite{he2024dataleakage2TAPE} have attempted to avoid the data leakage issue by collecting a new citation dataset, ensuring that the test papers are sampled from time periods after ChatGPT's data cut-off. However, these efforts remain limited to the citation domain, and the impact of graph structures in their datasets is not significant. Therefore, it's crucial to reconsider the methods employed to accurately evaluate the performance of LLMs on graph-related tasks. A fair, systematic, and comprehensive benchmark is also needed.
\vspace{-1em}
\subsection{Improving interpretability.} Interpretability, also known as explainability, refers to the ability to explain or present a model's behavior in terms that humans can understand~\cite{ijcai2024GFMSurvey}. In CDGL, most methods utilize Graph Neural Networks (GNNs). However, GNNs are typically treated as black boxes \cite{yuan2022GNNsBlackBoxAndExplainability} and lack interpretability. While these methods consider the relationships among various pieces of information in the graph during their design and have experimentally demonstrated effectiveness in CDGL, their improvements are often based on intuition and a high-level understanding of graph data. Consequently, their interpretability remains poor. To enhance interpretability, CDGL approaches that integrate Large Language Models (LLMs) have been proposed. This is primarily due to LLMs' reasoning and explanatory abilities, which enable them to produce user-friendly explanations for graph reasoning tasks \cite{ijcai2024GFMSurvey,chen2024graphwiz}. However, most of these approaches are limited by their reliance on TAGs; they can only provide explanations for conditional CDGL scenarios. Extending the powerful reasoning and explanatory abilities of LLMs from TAGs to non-TAGs is also an important direction. This expansion directly affects the use of CDGL models in applications with strict security requirements, such as network fault diagnosis~\cite{zhao2023communication1,ji2022communication2}.
\vspace{-1em}
\subsection{Unified Cross-domain Evaluation Metrics.}
As discussed in Sections~\ref{sec:cross-domain learning} and \ref{sec:cross-domain graph learning}, the ultimate goal of CDGL is to develop a foundational model capable of generalizing across any domain. However, most existing approaches evaluate model performance using classical graph-related task metrics, assessing cross-domain generalization ability based on performance across different datasets. This approach may not fully capture the model's ability to generalize from a global perspective. Consequently, designing a unified cross-domain evaluation metric to holistically assess generalization ability represents a promising direction for future research.
\vspace{-1em}
\subsection{Compatibility Analysis and Improvement across Domains.}
Existing CDGL approaches often treat data from different domains uniformly, aiming to uncover correlations in structure, features, or structural-feature patterns to enhance cross-domain graph representation learning for various tasks. However, in practice, the compatibility of information between domains varies significantly. Similar to how humans find it easier to transfer knowledge between closely related fields (e.g., applying mathematical principles to physics) than between unrelated fields (e.g., applying musical theory to physics), analyzing domain compatibility and making targeted adjustments during cross-domain learning is a critical area for further investigation.
\section{Conclusion}
Cross-domain graph learning (CDGL) has emerged as a prominent area of research towards really general-purpose graph foundation models in recent years. In this survey, we aim to deliver an in-depth overview of current strategies that empower graph models with cross-domain transferability. Firstly, we propose a comprehensive taxonomy spanning structure-, feature-, and mixture-oriented CDGL methods across three cross-domain scales (i.e., Limited, Conditional, and Open) and three cross-domain difficulty levels (i.e., High, Moderate, and Low). Secondly, we systematically review the representative studies according to the taxonomy. Finally, we discuss some remaining limitations and pinpoint several future research directions. Through this overview, we shed light on recent advances and challenges in CDGL, and we hope to catalyse further progress in the field.

\bibliography{main}
\bibliographystyle{IEEEtran}

\vspace{-2.5em}
\begin{IEEEbiography}[{\includegraphics[width=1in,height=1.25in,clip,keepaspectratio]{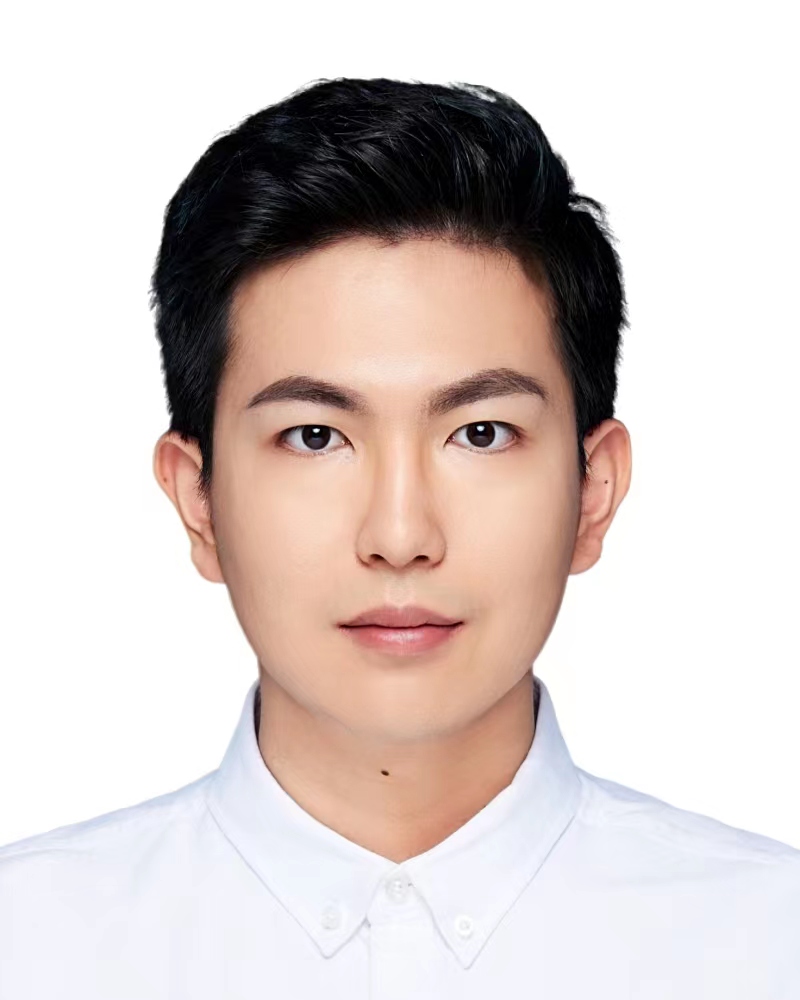}}]{Haihong Zhao} received his B.E. degree from The College of Software, Jilin University, Changchun, China, in 2019 and his M.sc degree from The School of Artificial Intelligence, Jilin University, Changchun, China, in 2022. He is currently a fall 2023 Ph.D. student at The Hong Kong University of Science and Technology (Guangzhou). His research interests include graph learning and data mining. He has published several papers as the leading author in top journals and conferences such as TMC, KDD, WWW, NeurIPS, and ICLR.
\end{IEEEbiography}


\begin{IEEEbiography}[{\includegraphics[width=1in,height=1.25in,clip,keepaspectratio]{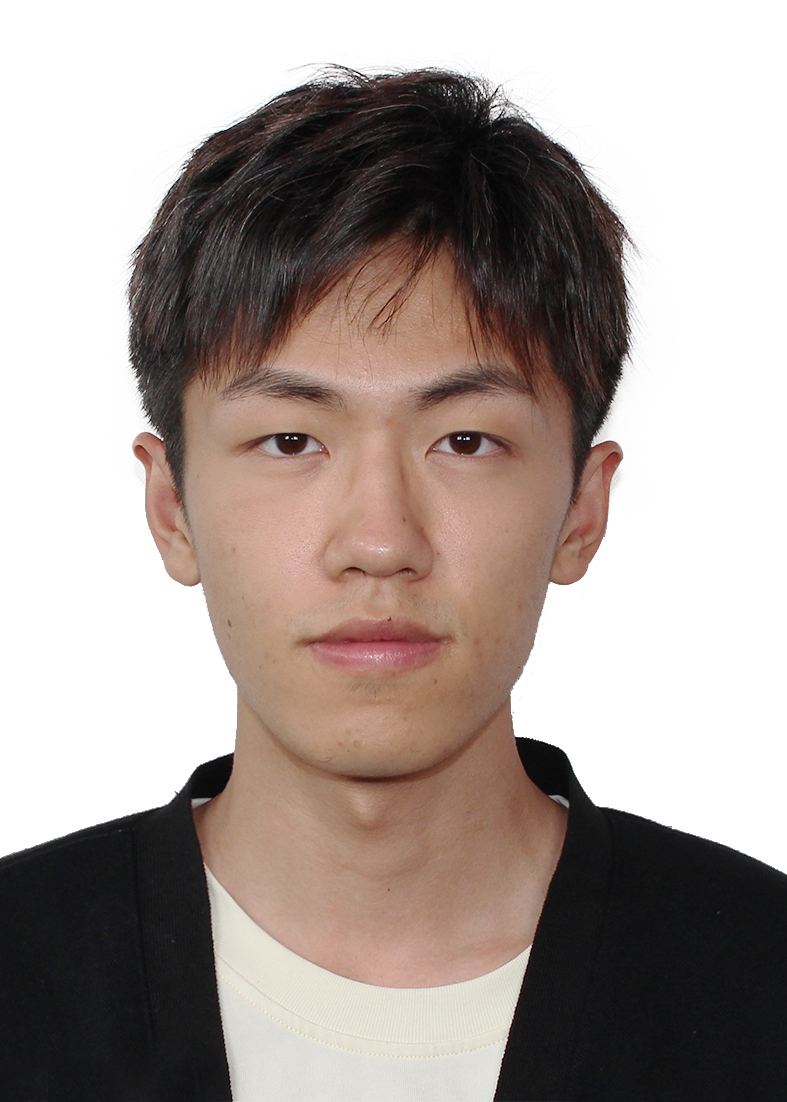}}]{Zhixun Li} received the B.Eng degree in Computer Science from Beijing Institute of Technology. He is currently a Ph.D. student in the Department of Systems Engineering and Engineering Management at the Chinese University of Hong Kong. His research interests mainly focus on deep graph learning and trustworthy AI. He has published papers at top international conferences, such as NeurIPS, SIGKDD, ICDM, and IJCAI. He has also reviewed papers in many top-tier conferences and journals, like SIGKDD, ICDM, ICDE, etc.
\end{IEEEbiography}


\begin{IEEEbiography}[{\includegraphics[width=1in,height=1.25in,clip,keepaspectratio]{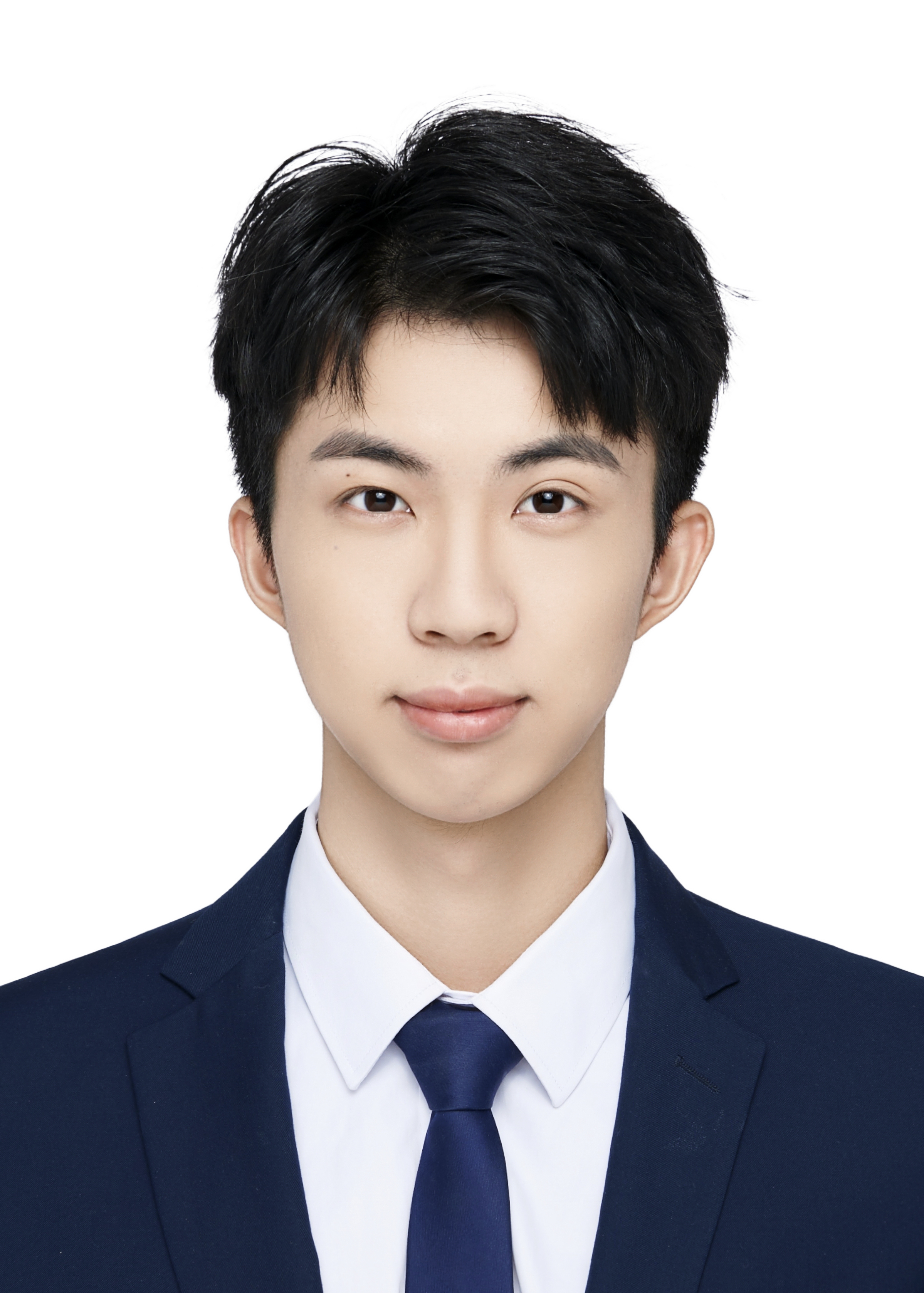}}]{Chenyi Zi} received his bachelor’s degree from Computer Science department, the South University of Technology. He is currently a fall 2023 MPhil. student at The Hong Kong University of Science and Technology (Guangzhou). His research interests include graph Pre-training, graph representation learning, and graph prompt learning.
\end{IEEEbiography}

\begin{IEEEbiography}[{\includegraphics[width=1in,height=1.25in,clip,keepaspectratio]{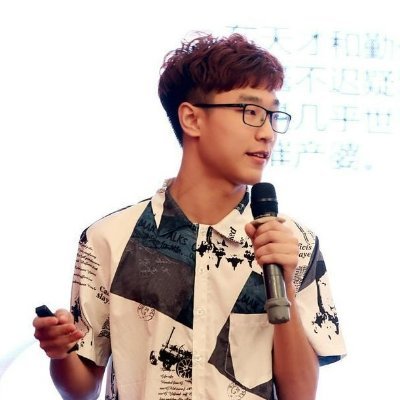}}]{Aochuan Chen} received his B.E. degree from The Department of Automation, Tsinghua University, Peking, China, in 2022. He is currently a spring 2024 Ph.D. student at The Hong Kong University of Science and Technology (Guangzhou). His research interests include scalable machine learning, graph learning, and LLMs. He has published several papers as the leading author in top conferences such as CVPR, ICLR, KDD, NeurIPS, and ICML.
\end{IEEEbiography}


\begin{IEEEbiography}[
{\includegraphics[width=1in,height=1.25in,clip,keepaspectratio]{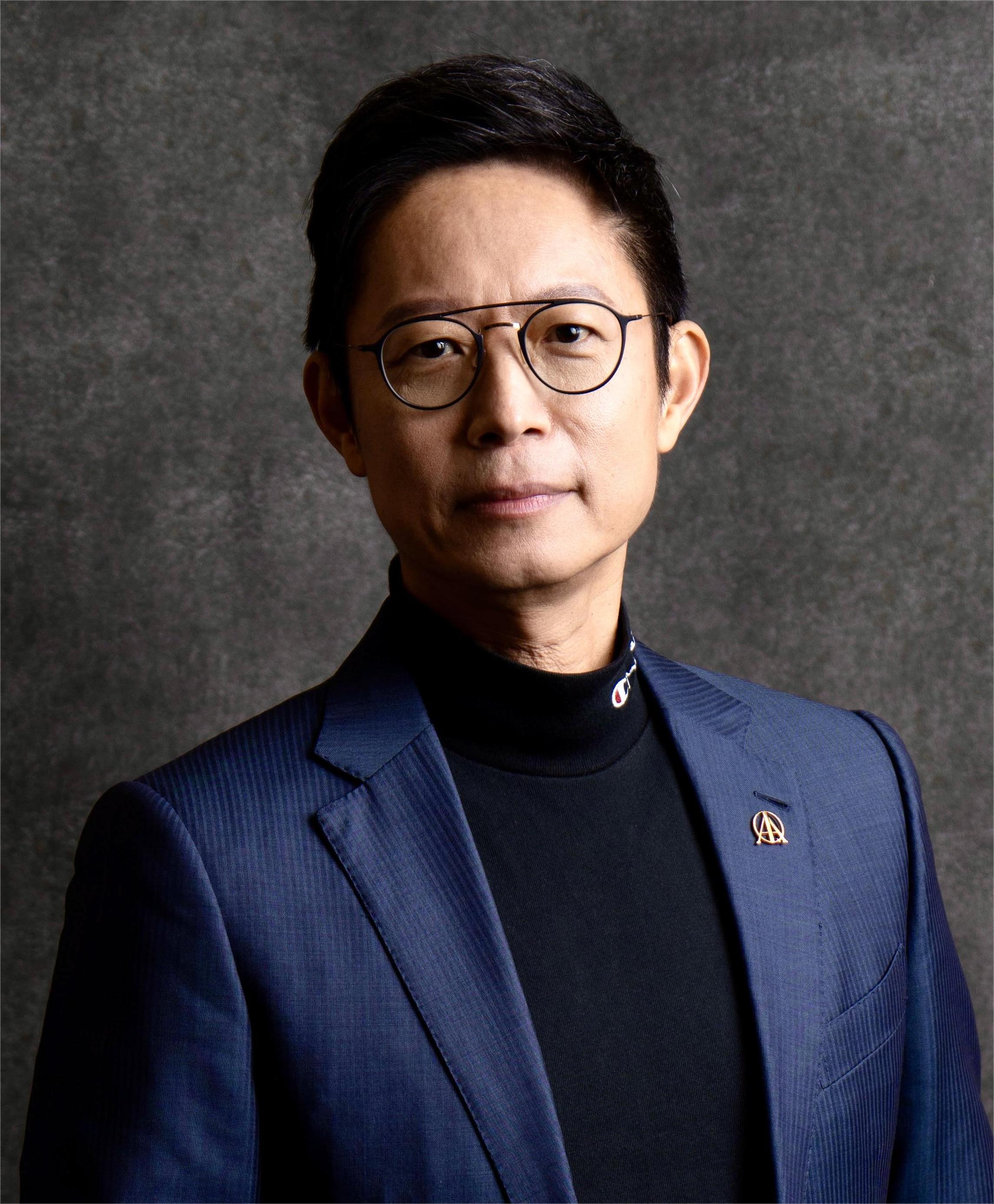}}
]{Fugee Tsung} (Senior Member, IEEE) received the Ph.D. degree from the University of Michigan, Ann Arbor, MI, USA, in 1997.

He is currently the Chair Professor with the Department of Industrial Engineering and Decision Analytics, The Hong Kong University of Science and Technology (HKUST), Hong Kong. He is also the Chair Professor with Data Science and Analytics Thrust and Computer Media and Arts Thrust, The Hong Kong University of Science and Technology (Guangzhou), Guangzhou, China. His research interests include quality analytics in advanced manufacturing and service processes, industrial big data, and statistical process control, monitoring, and diagnosis.

Prof. Tsung is a fellow of esteemed organizations such as ASA, ASQ, IISE, IAQ, and HKIE. He is a globally recognized expert in industrial analytics and quality engineering, listed among the top 2\% of most influential scientists worldwide by Stanford-Elsevier Mendeley Data in 2023 and 2024. As a Chair Professor at HKUST and HKUST (Guangzhou), he directs the Industrial and Intelligence Institute (Triple-I Institute) and the Quality and Data Analytics Laboratory (QLab). Most recently, he received the prestigious 2025 Shewhart Medal, awarded by ASQ and widely regarded as the “Nobel Prize of Quality Control.”

\end{IEEEbiography}
\vfill
\newpage 

\begin{IEEEbiography}[
{\includegraphics[width=1in,height=1.25in,clip,keepaspectratio]{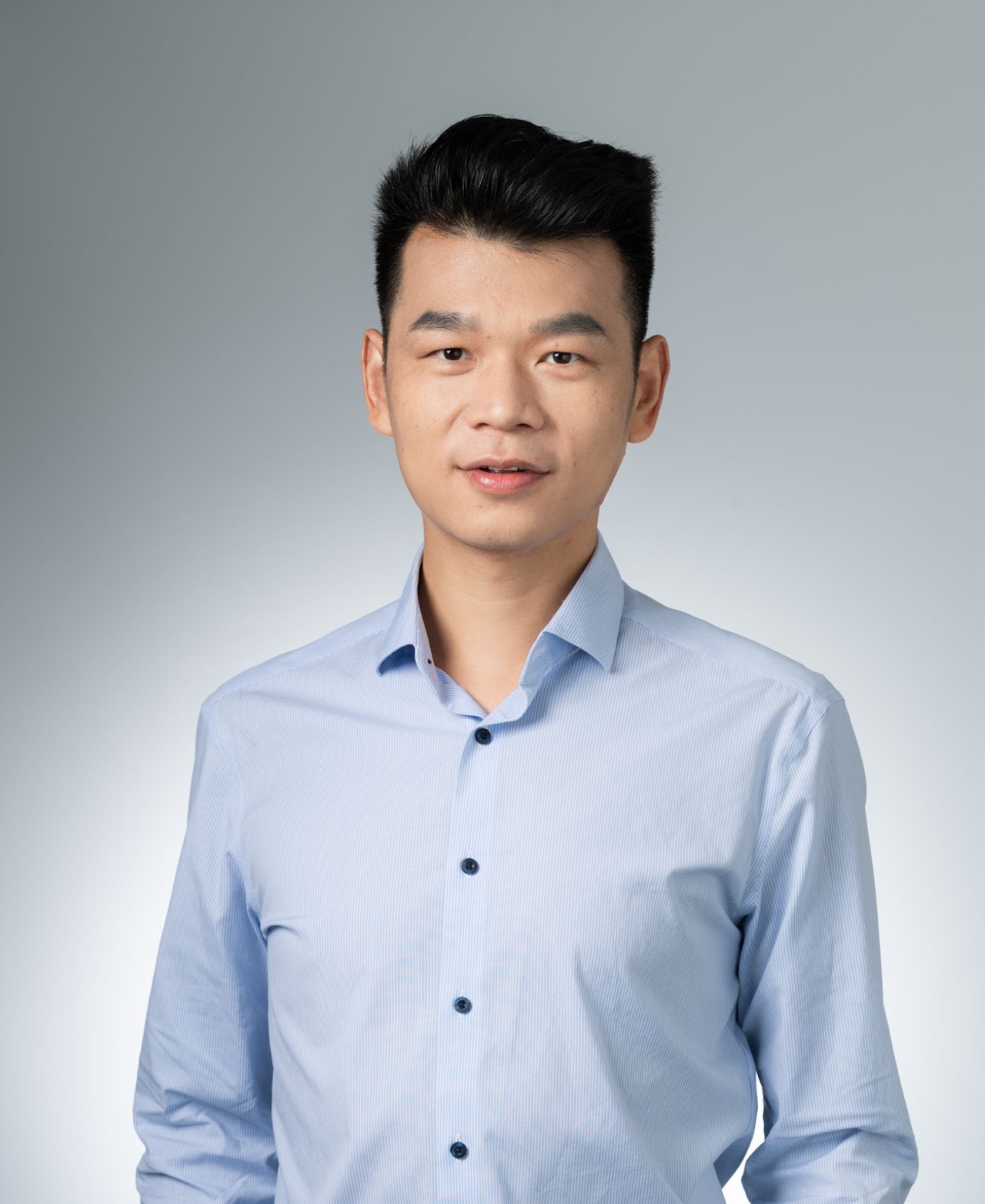}}
]{Jia Li} received the PhD degree from the Chinese University of Hong Kong, in 2021. He is an assistant professor with the Hong Kong University of Science and Technology (Guangzhou) and an affiliated assistant professor with the Hong Kong University
of Science and Technology. His research interests include machine learning, data mining, and deep graph learning. He has published several papers as the leading author in top conferences such as KDD, WWW and NeurIPS.
\end{IEEEbiography}


\begin{IEEEbiography}[
{\includegraphics[width=1in,height=1.25in,clip,keepaspectratio]{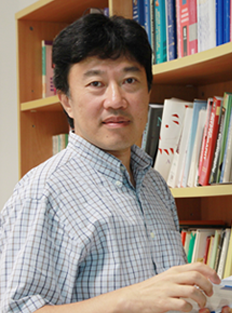}}
]{Jeffrey Xu YU} is a Professor of Data Science and Analytics Thrust, The Hong Kong University of Science and Technology (Guangzhou). His current main research interests include graph algorithms and systems, graph mining, vector databases, query processing, and optimization. Dr. Yu has served in over 300 organization and program committees of international conferences and workshops, such as the PC Co-Chair of PAKDD’10, DASFAA’11, ICDM’12, NDBC’13, CIKM’15 and CIKM’19, and as the Conference General Co-Chair of ICDM’18. He has also served as the Information Director and Executive Committee Member of ACM SIGMOD (2007–2011), Associate Editor of \textit{IEEE Transactions on Knowledge and Data Engineering} (2004–2008), Associate Editor of the \textit{VLDB Journal} (2007–2013), and Chair of the Steering Committee of the Asia Pacific Web Conference (2013–2016). He is currently serving as Associate Editor for \textit{ACM Transactions on Database Systems}, \textit{WWW Journal}, \textit{International Journal of Cooperative Information Systems}, \textit{Journal of Information Processing}, and \textit{Journal on Health Information Science and Systems}.
\end{IEEEbiography}

\vfill
\end{document}